\documentclass{article}

\PassOptionsToPackage{numbers, compress}{natbib}
\DeclareMathAlphabet{\mathcal}{OMS}{cmsy}{m}{n}
\usepackage{mathptmx}
\usepackage[final]{neurips_2018}
\usepackage[utf8]{inputenc} 
\usepackage{hyperref}       
\usepackage{url}            
\usepackage{booktabs}       
\usepackage{amsfonts}       
\usepackage{microtype}      
\usepackage{natbib}
\usepackage{amssymb}
\usepackage{graphicx}
\usepackage{subfigure}
\usepackage{amsmath}
\usepackage{algorithm}
\usepackage[noend]{algpseudocode}
\makeatletter
\def\BState{\State\hskip-\ALG@thistlm}

\usepackage{wrapfig}
\usepackage{xcolor}

\usepackage[utf8]{inputenc} 
\usepackage[T1]{fontenc}    
\usepackage{nicefrac}       
\usepackage{xcolor}
\usepackage[scaled=.8]{beramono} 
\usepackage{amsthm}
\usepackage{amsfonts}
\usepackage{bm}
\usepackage{array}
\usepackage{xspace}
\usepackage{pifont}
\usepackage{makecell}
\usepackage{xr}
\newif\ifcomments
\commentstrue

\ifcomments
	\newcommand{\sXX}[1]{\color{red}SO: (#1)\color{black}\xspace}
	\newcommand{\dXX}[1]{\color{olive}DK: (#1)\color{black}\xspace}
	\newcommand{\jXX}[1]{\color{orange}JH: (#1)\color{black}\xspace}
	\newcommand{\mXX}[1]{\color{cyan}ML: (#1)\color{black}\xspace}
	\newcommand{\gXX}[1]{\color{purple}GT: (#1)\color{black}\xspace}
	\newcommand{\mrXX}[1]{\color{blue}MR: (#1)\color{black}}
	\newcommand{\cXX}[1]{\color{olive}CA: (#1)\color{black}}
\else
	\newcommand{\sXX}[1]{}
	\newcommand{\dXX}[1]{}
	\newcommand{\jXX}[1]{}
	\newcommand{\mXX}[1]{}
	\newcommand{\gXX}[1]{}
	\newcommand{\mrXX}[1]{}
	\newcommand{\cXX}[1]{}
\fi

\makeatother

\title{Learning Abstract Options}

\author{
  Matthew Riemer, Miao Liu, and Gerald Tesauro \\
  IBM Research \\
  T.J. Watson Research Center,
  Yorktown Heights, NY \\
  \texttt{\{mdriemer, miao.liu1, gtesauro\}@us.ibm.com} \\
}

\begin{document}

\maketitle

\begin{abstract}

Building systems that autonomously create temporal abstractions from data is a key challenge in scaling learning and planning in reinforcement learning. One popular approach for addressing this challenge is the options framework \citep{Options}. However, only recently in \citep{OC} was a policy gradient theorem derived for online learning of general purpose options in an end to end fashion. In this work, we extend previous work on this topic that only focuses on learning a two-level hierarchy including options and primitive actions to enable learning simultaneously at multiple resolutions in time. We achieve this by considering an arbitrarily deep hierarchy of options where high level temporally extended options are composed of lower level options with finer resolutions in time. We extend results from \citep{OC} and derive policy gradient theorems for a deep hierarchy of options. Our proposed \textit{hierarchical option-critic} architecture is capable of learning internal policies, termination conditions, and hierarchical compositions over options without the need for any intrinsic rewards or subgoals.  Our empirical results in both discrete and continuous environments demonstrate the efficiency of our framework.
  
\end{abstract}

\section{Introduction}

In reinforcement learning (RL), \textit{options} \citep{Options,PrecupThesis} provide a general framework for defining temporally abstract courses of action for learning and planning. Extensive research has focused on discovering these temporal abstractions autonomously \citep{McgovernAndBarto,StolleAndPrecup,QCut,NIPS09,icml2012} while approaches that can be used in continuous state and/or action spaces have only recently became feasible \citep{konidaris,niekum,mann2015,asap,hDQN,macroactions,daniel2016}.
Most existing work has focused on finding subgoals (i.e. useful states for the agent) and then learning policies to achieve them. However, these approaches do not scale well because of their combinatorial nature. Recent work on \textit{option-critic} learning blurs the line between option discovery and option learning by providing policy gradient theorems for optimizing a two-level hierarchy of options and primitive actions \citep{OC}. These approaches have achieved success when applied to Q-learning on Atari games, but also in continuous action spaces \citep{oc_continuous} and with asynchronous parallelization \citep{Deliberation}. In this paper, we extend option-critic to a novel 
\textit{hierarchical option-critic} framework, 
presenting generalized policy gradient theorems that can be applied to an arbitrarily deep hierarchy of options. 

\begin{figure}[!h]
    \centering
    \vspace{-3mm}
    \includegraphics[scale=0.30]{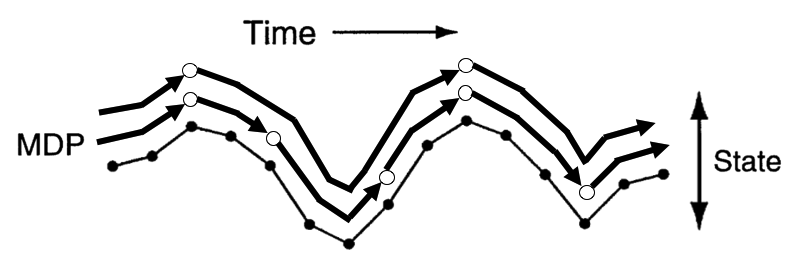}
    \vspace{-3mm}
    \caption{State trajectories over a three-level hierarchy of options. Open circles represent SMDP decision points while filled circles are primitive steps within an option. The low level options are temporally extended over primitive actions, and high level options are even further extended. }
    \label{HierarchyFigure}
    \vspace{-5mm}
\end{figure}

Work on learning with temporal abstraction is motivated by two key potential benefits over learning with primitive actions: long term credit assignment and exploration. Learning only at the primitive action level or even with low levels of abstraction slows down learning, because agents must learn longer sequences of actions to achieve the desired behavior. This frustrates the process of learning in environments with sparse rewards.  In contrast, 
agents that learn a high level decomposition of sub-tasks are able to explore the environment more effectively by exploring in the abstract action space rather than the primitive action space.  While the recently proposed deliberation cost \citep{Deliberation} can be used as a margin that effectively controls how temporally extended options are, the standard two-level version of the option-critic framework is still ill-equipped to learn complex tasks that require sub-task decomposition at multiple quite different temporal resolutions of abstraction. In Figure \ref{HierarchyFigure} we depict how we overcome this obstacle
to learn a deep hierarchy of options. The standard two-level option hierarchy constructs a Semi-Markov Decision Process (SMDP), where new options are chosen when temporally extended sequences of primitive actions are terminated. In our framework, we consider not just options and primitive actions, but also an arbitrarily deep hierarchy of lower level and higher level options. Higher level options represent a further temporally extended SMDP than the low level options below as they only have an opportunity to terminate when all lower level options terminate. 

We will start by reviewing related research and by describing the seminal work we build upon 
in this paper that first derived policy gradient theorems for learning with primitive actions \citep{Sutton2000} and options \citep{OC}.  We will then describe the core ideas of our approach, presenting hierarchical intra-option policy and termination gradient theorems. We leverage this new type of policy gradient learning to construct a \textit{hierarchical option-critic} architecture, which generalizes the \textit{option-critic} architecture to an arbitrarily deep hierarchy of options. Finally, we demonstrate the empirical benefit of this architecture over standard \textit{option-critic} when applied to RL benchmarks. To the best of our knowledge, this is the first general purpose end-to-end approach for learning a deep hierarchy of options beyond two-levels in RL settings, scaling to very large domains at comparable efficiency. 

\vspace{-2mm}
\section{Related Work}

Our work is related to recent literature on learning to compose skills in RL. As an example, \citet{composeskills} leverages a logic for combining pre-learned skills by learning an embedding to represent the combination of a skill and state. Unfortunately, their system relies on a pre-specified sub-task decomposition into skills. In \citep{languagehrl}, the authors propose to ground all goals in a natural language description space. Created descriptions can then be high level and express a sequence of goals. While these are interesting directions for further exploration, we will focus on a more general setting without provided natural language goal descriptions or sub-task decomposition information. 

Our work is also related to methods that learn to decompose the problem over long time horizons. A prominent paradigm for this is Feudal Reinforcement Learning \citep{feudalrl}, which learns using manager and worker models. Theoretically, this can be extended to a deep hierarchy of managers and their managers as done in the original work for a hand designed decomposition of the state space. Much more recently, \citet{fun} showed the ability to successfully train a Feudal model end to end with deep neural networks for the Atari games. However, this has only been achieved for a two-level hierarchy (i.e. one manager and one worker). We can think of Feudal approaches as learning to decompose the problem with respect to the state space, while the options framework learns a temporal decomposition of the problem. Recent work \citep{hac} also breaks down the problem over a temporal hierarchy, but like \citep{fun} is based on learning a latent goal representation that modulates the policy behavior as opposed to options. Conceptually, options stress choosing among skill abstractions and Feudal approaches stress the achievement of certain kinds of states. Humans tend to use both of these kinds of reasoning when appropriate and we conjecture that a hybrid approach will likely win out in the end. Unfortunately, in the space available we feel that we cannot come to definitive conclusions about the precise nature of the differences and potential synergies of these approaches.  

The concept of learning a hierarchy of options is not new. It is an obviously desirable extension of options envisioned in the original papers. However, actually learning a deep hierarchy of options end to end has received surprisingly little attention to date. Compositional planning where options select other options was first considered in \citep{icml2012}. The authors provided a generalization of value iteration to option models for multiple subgoals, leveraging explicit subgoals for options. Recently, \citet{ddo} successfully trained a hierarchy of options end to end for imitation learning. Their approach leverages an EM based algorithm for recursive discovery of additional levels of the option hierarchy. Unfortunately, their approach is only applicable to imitation learning and not general purpose RL.  
We are the first to propose theorems along with a practical algorithm and architecture to train arbitrarily deep hierarchies of options end to end using policy gradients, maximizing the expected return. 

\section{Problem Setting and Notation} 

A Markov Decision Process (MDP) is defined with a set of states $\mathcal{S}$, a set of actions $\mathcal{A}$, a transition function $\mathcal{P}: \mathcal{S} \times \mathcal{A} \rightarrow ( \mathcal{S} \rightarrow [0,1])$ and a reward function $r: \mathcal{S} \times \mathcal{A} \rightarrow \mathbb{R}$. We follow \citep{OC} and  develop  our  ideas  assuming  discrete  state  and  action sets, while our results extend to continuous spaces using usual measure-theoretic assumptions as demonstrated in our experiments. A policy is defined as a probability distribution over actions conditioned on states, $\pi: \mathcal{S} \rightarrow \mathcal{A} \rightarrow [0,1]$. The value function of a policy $\pi$ is the expected return $V_\pi(s) = \mathbb{E}_\pi[\sum_{t=0}^{\infty} \gamma^t r_{t+1}|s_0=s]$ with an action-value function of $Q_\pi(s,a) = \mathbb{E}_\pi[\sum_{t=0}^{\infty} \gamma^t r_{t+1}|s_0=s,a_0=a]$ where $\gamma \in [0,1)$ is the \emph{discount factor}.

\textbf{Policy gradient methods} \citep{Sutton2000,Konda2000} consider  the  problem  of  improving a  policy  by  performing  stochastic  gradient  descent to  optimize  a  performance  objective  over a family  of parametrized  stochastic  policies, $\pi_\theta$. The policy gradient  theorem  \citep{Sutton2000}  provides  the  gradient  of the  discounted  reward  objective  with  respect  to $\theta$ in a straightforward expression. The  objective  is  defined  with  respect  to a  designated  starting  state $s_0: \rho(\theta,s_0) = \mathbb{E}_{\pi_\theta}[\sum_{t=0}^{\infty} \gamma^t r_{t+1}|s_0]$. The policy gradient theorem shows that: $\frac{\partial \rho(\theta,s_0)}{\partial \theta} = \sum_s \mu_{\pi_\theta}(s|s_0) \sum_a \frac{\partial \pi_\theta(a|s)}{\partial \theta} Q_{\pi_\theta}(s,a)$, where $\mu_{\pi_\theta}(s|s_0) = \sum_{t=0}^{\infty} \gamma^t P(s_t=s|s_0)$ is the  discounted weighting of the states along the trajectories starting from initial state $s_0$. 

\textbf{The options framework} \citep{Options,PrecupThesis}  provides formalism for  the  idea  of  temporally  extended  actions.  A Markovian option $o \in \Omega$ is  a  triple $(I_o,\pi_o,\beta_o)$ where $I_o \subseteq S$ represents an initiation set, $\pi_o$ represents an intra-option policy,  and
$\beta_o: \mathcal{S} \rightarrow [0,1]$ represents  a  termination  function.  Like most option discovery algorithms, we  assume  that all options are available everywhere. MDPs  with options become SMDPs \citep{markov1994} with an associated optimal value function over options $V^*_\Omega(s)$ and option-value function $Q^*_\Omega(s,o)$  \citep{Options,PrecupThesis}. 

\textbf{The option-critic architecture} \citep{OC} utilizes a call-and-return option execution model. An agent picks option $o$ according to its policy over options $\pi_\Omega(o|s)$, then follows the intra-option policy $\pi(a|s,o)$ until termination (as determined by $\beta(s,o)$), which triggers a repetition of this procedure. Let $\pi_{\theta}(a|s,o)$ denote the intra-option policy of option $o$ parametrized by $\theta$ and $\beta_{\phi}(s,o)$ the termination function of $o$ parameterized by $\phi$. Like policy gradient methods, the option-critic architecture optimizes directly for the discounted return  expected  over  trajectories  starting  at  a  designated  state $s_0$ and  option $o_0$:  $\rho(\Omega,\theta,\phi,s_0,o_0) = \mathbb{E}_{\Omega,\pi_\theta,\beta_\phi} [\sum_{t=0}^{\infty} \gamma^t r_{t+1}|s_0,o_0]$. 
The option-value function is then:

\vspace{-4mm}
\begin{equation}
Q_\Omega(s,o) = \sum_a \pi_{\theta}(a|s,o)Q_U(s,o,a),
\end{equation}
\vspace{-4mm}

where $Q_U: \mathcal{S} \times \Omega \times \mathcal{A} \rightarrow \mathbb{R}$ is the value of selecting an action given the context of a state-option pair:

\vspace{-5mm}
\begin{equation}
Q_U(s,o,a) = r(s,a) + \gamma \sum_{s'} P(s'|s,a)U(s',o).
\end{equation}
\vspace{-4mm}

The $(s,o)$ pairs define an augmented state space \citep{Levy2011}. The option-critic architecture instead leverages the function $U: \Omega \times \mathcal{S} \rightarrow \mathbb{R}$ which is called the option-value function upon arrival \citep{Options}. The value of selecting option $o$ upon entering state $s'$ is:

\vspace{-5mm}
\begin{equation} \label{U2}
U(s',o) = ( 1 - \beta_{\phi}(s',o) ) Q_\Omega(s',o) + \beta_{\phi}(s',o) V_\Omega(s').
\end{equation}
\vspace{-4mm}

For notation clarity, we omit  $\theta$ and $\phi$ which $Q_U$ and $U$ both  depend  on. The intra-option policy gradient theorem results from taking the derivative of the expected discounted return with respect to the intra-option policy parameters $\theta$ and defines the update rule for the intra-option policy:

\vspace{-4mm}
\begin{equation}
\frac{\partial Q_\Omega(s_0,o_0)}{\partial \theta} = \sum_{s,o} \mu_\Omega(s,o|s_0,o_0) \sum_a \frac{\partial \pi_{\theta}(a|s,o)}{\partial \theta} Q_U(s,o,a).
\end{equation}
\vspace{-4mm}

where $\mu_\Omega$ is the discounted weighting of $(s,o)$ along trajectories originating from
$(s_0,o_0): \mu_\Omega(s,o|s_0,o_0) = \sum_{t=0}^{\infty} \gamma^t P(s_t=s,o_t=o|s_0,o_0)$. The termination gradient theorem results from taking the derivative of the expected discounted return with respect to the termination policy parameters $\phi$ and defines the update rule for the termination policy for the initial condition $(s_1,o_0)$:

\vspace{-4mm}
\begin{equation}
\frac{\partial Q_\Omega(s,o)}{\partial \phi} = \sum_a \pi_{\theta}(a|s,o) \sum_{s'} \gamma P(s'|s,a) \frac{\partial U(s',o)}{\partial \phi},
\end{equation}
\vspace{-3mm}

\vspace{-4mm}
\begin{equation}
\frac{\partial U(s_1,o_0)}{\partial \phi} = - \sum_{s',o} \mu_\Omega(s',o|s_1,o_o) \frac{\partial \beta_{\phi}(s',o)}{\partial \phi} A_\Omega(s',o),
\end{equation}
\vspace{-4mm}

where $\mu_\Omega$ is now the discounted weighting of $(s,o)$ from $(s_1,o_0): \mu_\Omega(s,o|s_1,o_0) = \sum_{t=0}^\infty \gamma^t P(s_{t+1}=s, o_t=o | s_1,o_0)$. $A_\Omega$ is the advantage function over options: $A_\Omega(s',o) = Q_\Omega(s',o) - V_\Omega(s')$. 

\vspace{-3mm}
\section{Learning Options with Arbitrary Levels of Abstraction}
\vspace{-2mm}

\textbf{Notation:} As it makes our equations much clearer and more condensed we adopt the notation $x^{i:i+j} = x^i, ..., x^{i+j}$. This implies that  $x^{i:i+j}$ denotes a list of variables in the range of $i$ through $i+j$.

\begin{wrapfigure}[18]{r}{0.65\textwidth}
\vspace{-5mm}
\begin{center}
	\includegraphics[scale=0.14]{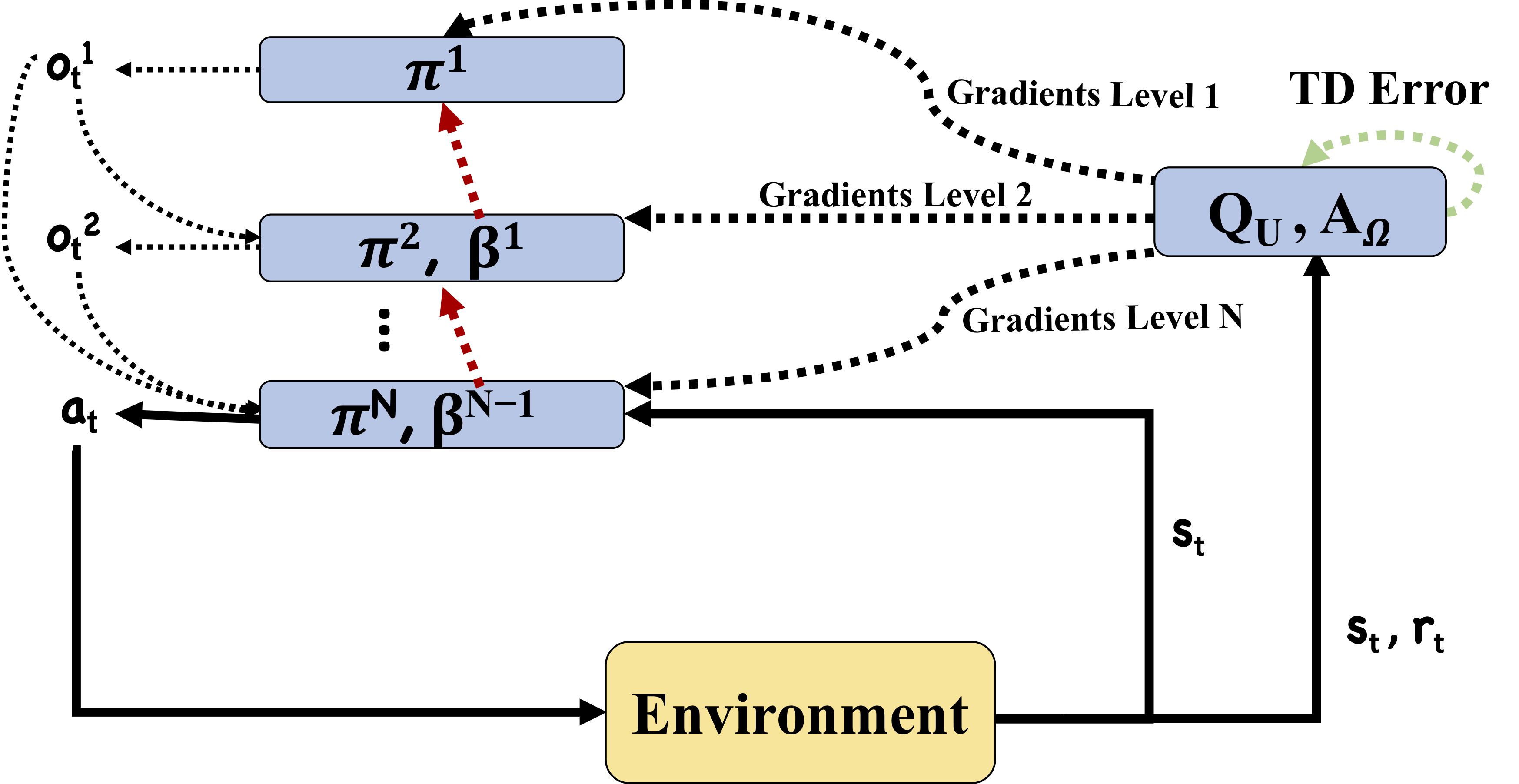}
\end{center}
\vspace{-3mm}
\caption{A diagram describing our proposed hierarchical option-critic architecture. Dotted lines represent processes within the agent while solid lines represent processes within the environment. Option selection is top down through the hierarchy and option termination is bottom up (represented with red dotted lines).}
	\label{Architecture}
\end{wrapfigure}

\textbf{The hierarchical options framework} that we introduce in this work considers an agent that learns using an $N$ level hierarchy of policies, termination functions, and value functions. Our goal is to extend the ideas of the option-critic architecture in such a way that our framework simplifies to policy gradient based learning when $N=1$ and option-critic learning when $N=2$. At each hierarchical level above the lowest primitive action level policy, we consider an available set of options $\Omega^{1:N-1}$ that is a subset of the total set of available options $\Omega$. This way we keep our view of the possible available options at each level very broad. On one extreme, each hierarchical level may get its own unique set of options and on the other extreme each hierarchical level may share the same set of options. We present a diagram of our proposed architecture in Figure \ref{Architecture}. 


We denote $\pi_{\theta^1}^1(o^1|s)$ as the policy over the most abstract options in the hierarchy $o^1 \in \Omega_1$ given the state $s$. For example, $\pi^1=\pi_\Omega$ from our discussion of the option-critic architecture. Once $o^1$ is chosen with policy $\pi^1$, then we go to policy $\pi_{\theta^2}^2(o^2|s,o^1)$, which is the next highest level policy, to select $o^2 \in \Omega_2$ conditioning it on both the current state $s$ and the selected highest level option $o^1$. This process continues on in the same fashion stepping down to policies at lower levels of abstraction conditioned on the augmented state space considering all selected higher level options until we reach policy $\pi_{\theta^N}^N(a|s,o^{1:N-1})$. $\pi^N$ is the lowest level policy and it finally selects over the primitive action space conditioned on all of the selected options.  

Each level of the option hierarchy has a complimentary termination function $\beta_{\phi^1}^1(s,o^1),...,\beta_{\phi^{N-1}}^{N-1}(s,o^{1:N-1})$ that governs the termination pattern of the selected option at that level. We adopt a bottom up termination strategy where high level options only have an opportunity to terminate when all of the lower level options have terminated first. For example, $o_t^{N-2}$ cannot terminate until $o_t^{N-1}$ terminates at which point we can assess $\beta_{\phi^{N-2}}^{N-2}(s,o^{1:N-2})$ to see whether $o_t^{N-2}$ terminates. If it did terminate, this would allow $o_t^{N-3}$ the opportunity to asses if it should terminate and so on. This condition ensures that higher level options will be more temporally extended than their lower level option building blocks, which is a key motivation of this work. 

The final key component of our system is the value function over the augmented state space at each level of abstraction. To enable comprehensive reasoning about the policies at each level of the option hierarchy, we need to maintain value functions that consider the state and every possible combination of active options and actions $V_\Omega(s),Q_\Omega(s,o^1),...,Q_\Omega(s,o^{1:N-1},a)$. These value functions collectively serve as the \textit{critic} in our analogy to the actor-critic and option-critic training paradigms. 

\subsection{Generalizing the Option Value Function to N Hierarchical Levels}

Like policy gradient methods and option-critic, the hierarchical options framework optimizes directly for the discounted return  expected  over  all trajectories  starting  at  a  state $s_0$ with active options $o_0^{1:N-1}$:

\vspace{-5mm}
\begin{equation}
\begin{split}
\rho(\Omega^{1:N-1},\theta^{1:N},\phi^{1:N-1},s_0,o_0^{1:N-1}) =  \mathbb{E}_{\Omega^{1:N-1}, \pi^{1:N}_\theta, \beta^{1:N-1}_\phi} [\sum_{t=0}^{\infty} \gamma^t r_{t+1}|s_0,o_0^{1:N-1}]
\end{split}
\end{equation}
\vspace{-5mm}

This return depends on the policies and termination functions at each level of abstraction.  We now consider the option value function for understanding reasoning about an option $o^\ell$ at level $1 \leq \ell \leq N$ based on the augmented state space $(s,o^{1:\ell-1})$:

\vspace{-4mm}
\begin{equation} \label{PGQ}
Q_\Omega(s,o^{1:\ell-1}) = \sum_{o^\ell} \pi^\ell_{\theta^\ell}(o^\ell|s,o^{1:\ell-1})Q_\Omega(s,o^{1:\ell})
\end{equation}
\vspace{-5mm}

Note that in order to simplify our notation we write $o^\ell$ as referring to both abstract and primitive actions. As a result, $o^N$ is equivalent to $a$, leveraging the primitive action space $\mathcal{A}$. Extending the meaning of $Q_U$ from \citep{OC}, we define the corresponding value of executing an option in the presence of the currently active higher level options by integrating out the lower level options: 
\begin{equation} \label{QwithU}
\begin{split}
Q_U(s,o^{1:\ell})\!=\!\sum_{o^N}\!\!...\!\!\sum_{o^{\ell+1}}\!\prod_{j=\ell+1}^N \!\!\!\!\pi^j(o^j|s,o^{1:j-\!1})[r(s,\!o^N)\!+\!\gamma\!\sum_{s'}\!P(s'|s,o^N) U(s',o^{1:\ell})].
\end{split}
\end{equation}
The hierarchical option value function upon arrival $U$ with augmented state $(s,o^{1:\ell})$ is defined as:
\begin{equation} \label{Ueq}
\begin{split}
U(s',o^{1:\ell}) = \underbrace{(1-\beta_{\phi^{N-1}}^{N-1}(s',o^{1:N-1}))Q_\Omega(s',o^{1:N-1})}_{\text{none terminate ($N \geq 1$)}} +  \underbrace{V_\Omega(s') \prod_{j=N-1}^{1} \beta_{\phi^j}^j(s',o^{1:j})}_{\text{all options terminate ($N \geq 2$)}}  + \\
\underbrace{ Q_\Omega(s',o^{1:\ell}) \sum_{q=N-1}^{\ell+1} (1-\beta_{\phi^{q-1}}^{q-1}(s',o^{1:q-1})) \prod_{z=N-1}^q \beta_{\phi^z}^z(s',o^{1:z})}_{\text{only lower level options terminate ($N \geq 3$)}} + \\
\underbrace{ \sum_{i=1}^{\ell - 1}(1-\beta_{\phi^i}^i(s',o^{1:i})) Q_\Omega(s',o^{1:i}) \prod_{k=i+1}^{N-1} \beta_{\phi^k}^k(s',o^{1:k}) }_{\text{some relevant higher level options terminate ($N \geq 3$)}}].
\end{split}
\end{equation}
We explain the derivation of this equation \footnote{Note that when no options terminate, as in the first term in equation \eqref{Ueq}, the lowest level option does not terminate and thus no higher level options have the opportunity to terminate.} in the Appendix \ref{Proof:0}.
Finally, before we can extend the policy gradient theorem, we must establish the Markov chain along which we can measure performance for options with $N$ levels of abstraction. This is derived in the Appendix \ref{Proof:1}.

\subsection{Generalizing the Intra-option Policy Gradient Theorem}

We can think of actor-critic architectures, generalizing to the option-critic architecture as well, as pairing a critic with each actor network so that the critic has additional information about the value of the actor's actions that can be used to improve the actor's learning. However, this is derived by taking gradients with respect to the parameters of the policy while optimizing for the expected discounted return. The discounted return is approximated by a critic (i.e. value function) with the same augmented state-space as the policy being optimized for. As examples, an actor-critic policy $\pi(a|s)$ is optimized by taking the derivative of its parameters with respect to $V_\pi(s)$ \citep{Sutton2000} and an option-critic policy $\pi(a|s,o)$ is optimized by taking the derivative of its parameters with respect to $Q_\Omega(s,o)$ \citep{OC}. The intra-option policy gradient theorem \citep{OC} is an important contribution, outlining how to optimize for a policy that is also associated with a termination function. As the policy over options in that work never terminates, it does not need a special training methodology and the option-critic architecture allows the practitioner to pick their own method of learning the policy over options while using Q Learning as an example in their experiments. We do the same for our highest level policy $\pi^1$ that also never terminates. For all other policies $\pi^{2:N}$ we perform a generalization of actor-critic learning by providing a critic at each level and guiding gradients using the appropriate critic. 

We now seek to generalize the intra-option policy gradients theorem, deriving the update rule for a policy at an arbitrary level of abstraction $\pi^\ell$ by taking the gradient with respect to $\theta^\ell$ using the value function with the same augmented state space $Q_\Omega(s,o^{1:\ell-1})$. Substituting from equation \eqref{PGQ} we find:

\vspace{-4mm}
\begin{equation}
\frac{\partial Q_\Omega(s,o^{1:\ell-1})}{\partial \theta^\ell} = \frac{\partial}{\partial \theta^\ell} \sum_{o^\ell} \pi^\ell_{\theta^\ell}(o^\ell|s,o^{1:\ell-1})Q_U(s,o^{1:\ell}).
\end{equation}
\vspace{-4mm}

\textbf{Theorem 1} (Hierarchical Intra-option Policy Gradient Theorem). \textit{Given an $N$ level hierarchical set  of  Markov  options  with  stochastic  intra-option  policies differentiable in their parameters $\theta^\ell$ governing each policy $\pi^\ell$, the gradient of the expected discounted return with respect to $\theta^\ell$ and initial conditions $(s_0,o_0^{1:N-1})$ is:}
\vspace{-3mm}
\[ \sum_{s,o^{1:\ell-1}} \mu_\Omega(s,o^{1:\ell-1}|s_0,o_0^{1:\ell-1}) \sum_{o^\ell} \frac{\partial \pi^\ell_{\theta^\ell}(o^\ell|s,o^{1:\ell-1})}{\partial \theta^\ell} Q_U(s,o^{1:\ell}),\]
where $\mu_\Omega$ is a discounted weighting of augmented state tuples along trajectories starting from $(s_0,o_0^{1:N-1}): \mu_\Omega(s,o^{1:\ell-1}|s_0,o_0^{1:\ell-1}) = \sum_{t=0}^\infty \gamma^t P(s_t = s, o_t^{1:\ell-1} = o^{1:\ell-1}|s_0,o_0^{1:\ell-1})$. A proof is in Appendix \ref{Proof:2}. 
\vspace{-2mm}
\subsection{Generalizing the Termination Gradient Theorem}
\vspace{-2mm}
We now turn our attention to computing gradients for the termination functions $\beta^\ell$ at each level, assumed to be stochastic and
differentiable with respect to the associated parameters $\phi^\ell$.

\vspace{-4mm}
\begin{equation} \label{simplederivative}
\frac{\partial Q_\Omega(s,o^{1:\ell})}{\partial \phi^\ell} = \sum_{o^N}\!\!...\!\!\sum_{o^{\ell+1}}\!\prod_{j=\ell+1}^N \pi^j(o^j|s,o^{1:j-\!1})\!\gamma\!\sum_{s'}\!P(s'|s,o^N) \frac{\partial U(s',o^{1:\ell})}{\partial \phi^\ell}
\end{equation}

Hence, the key quantity is the gradient of $U$. This is a natural consequence of call-and-return execution, where termination function quality can only be evaluated upon entering the next state.

\textbf{Theorem 2} (Hierarchical Termination Gradient Theorem). \textit{Given an $N$ level hierarchical set  of  Markov  options  with  stochastic  termination  functions differentiable in their parameters $\phi^\ell$ governing each function $\beta^\ell$, the gradient of the expected discounted return with respect to $\phi^\ell$ and initial conditions $(s_1,o_0^{1:N-1})$ is:}
\vspace{-3mm}
\[ -\sum_{s,o^{1:\ell}}  \mu_\Omega(s,o^{1:N-1}|s_1,o_0^{1:N-1}) \prod_{i=\ell+1}^{N-1} \beta_{\phi^i}^i(s,o^{1:i}) \frac{\partial \beta_{\phi^\ell}^\ell(s,o^{1:\ell})}{\partial \phi^\ell} A_\Omega(s,o^{1:\ell}) \]
where $\mu_\Omega$  is a discounted weighting of augmented state tuples along trajectories starting from $(s_1,o_0^{1:N-1}): \mu_\Omega(s,o^{1:\ell-1}|s_1,o_0^{1:N-1}) = \sum_{t=0}^\infty \gamma^t P(s_t = s, o_t^{1:\ell} = o^{1:\ell}|s_1,o_0^{1:N-1})$. $A_\Omega$ is the generalized advantage function over a hierarchical set of options $A_\Omega(s',o^{1:\ell}) = Q_\Omega(s',o^{1:\ell}) - V_\Omega(s) [\prod_{j=\ell-1}^{1} \beta_{\phi^j}^j(s',o^{1:j})]  - \sum_{i=1}^{\ell - 1}(1-\beta_{\phi^i}^i(s',o^{1:i})) Q_\Omega(s',o^{1:i}) [\prod_{k=i+1}^{\ell - 1} \beta_{\phi^k}^k(s',o^{1:k})]$. $A_\Omega$ compares the advantage of not terminating the current option with a probability weighted expectation based on the likelihood that higher level options also terminate. In \citep{OC} this expression was simple as there was not a hierarchy of higher level termination functions to consider. A proof is in Appendix \ref{Proof:3}.

It is interesting to see the emergence of an advantage function as a natural consequence of the derivation. As in \citep{OC} where this kind of relationship also appears, the advantage function gives the theorem an intuitive interpretation. When the option choice is sub-optimal at level $\ell$ with respect to the expected value of terminating option $\ell$, the advantage function is negative and increases the odds of terminating that option. A new concept, not paralleled in the option-critic derivation, is the inclusion of a $\prod_{i=\ell+1}^{N-1} \beta_{\phi^i}^i(s,o^{1:i})$ multiplicative factor. This can be interpreted as discounting gradients by the likelihood of this termination function being assessed as $\beta^\ell$ is only used if all lower level options terminate. This is a natural consequence of multi-level call-and-return execution. 

\section{Experiments}

We would now like to empirically validate the efficacy of our proposed hierarchical option-critic (HOC) model. We achieve this by exploring benchmarks in the tabular and non-linear function approximation settings. In each case we implement an agent that is restricted to primitive actions (i.e. $N=1$), an agent that leverages the option-critic (OC) architecture (i.e. $N=2$), and an agent with the HOC architecture at level of abstraction $N=3$. We will demonstrate that complex RL problems may be more easily learned using beyond two levels of abstraction and that the HOC architecture can successfully facilitate this level of learning using data from scratch.   

\begin{wrapfigure}{r}{0.55\textwidth}
 \vspace{-7mm}
\hspace{-2mm}
\includegraphics[scale=0.16]{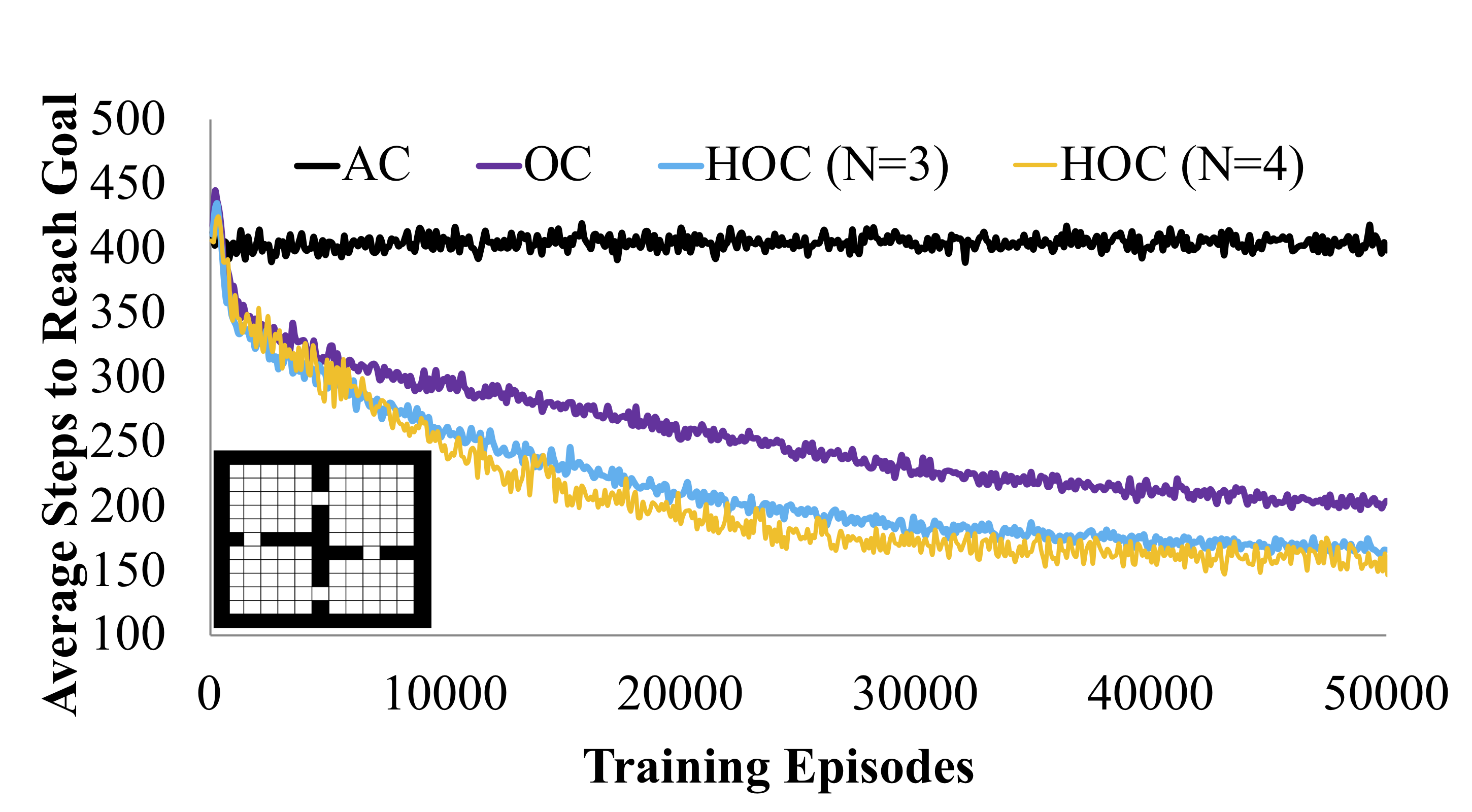}
\vspace{-7mm}
\caption{Learning performance as a function of the abstraction level for a nonstationary four rooms domain where the goal location changes every episode.}
\label{Fourrooms}
\vspace{-3mm}
\end{wrapfigure}

For our tabular architectures, we followed protocol from \citep{OC} and chose to parametrize the intra-option policies with softmax distributions and the terminations with sigmoid functions. The policy over options was learned using intra-option Q-learning. We also implemented primitive actor-critic (AC) using a softmax policy. For the non-linear function approximation setting, we trained our agents using A3C \citep{A3C}. Our primitive action agents conduct A3C training using a convolutional network when there is image input followed by an LSTM to contextualize the state. This way we ensure that benefits seen from options are orthogonal to those seen from these common neural network building blocks. We follow \citep{Deliberation} to extend A3C to the Asynchronous Advantage Option-Critic (A2OC) and Asynchronous Advantage Hierarchical Option-Critic architectures (A2HOC). We include detailed algorithm descriptions for all of our experiments in Appendix \ref{ExpApp}. 
We also conducted hyperparameter optimization that is summarized along with detail on experimental protocol in Appendix \ref{ExpApp}. 
In all of our experiments, we made sure that the two-level OC architecture had access to more total options than the three level alternative and that the three level architecture did not include any additional hyperparameters. This ensures that empirical gains are the result of increasingly abstract options. 

\vspace{-3mm}
\subsection{Tabular Learning Challenge Problems}
\vspace{-2mm}

\begin{wrapfigure}{r}{0.45\textwidth}
\vspace{-3mm}
	\includegraphics[scale=0.125]{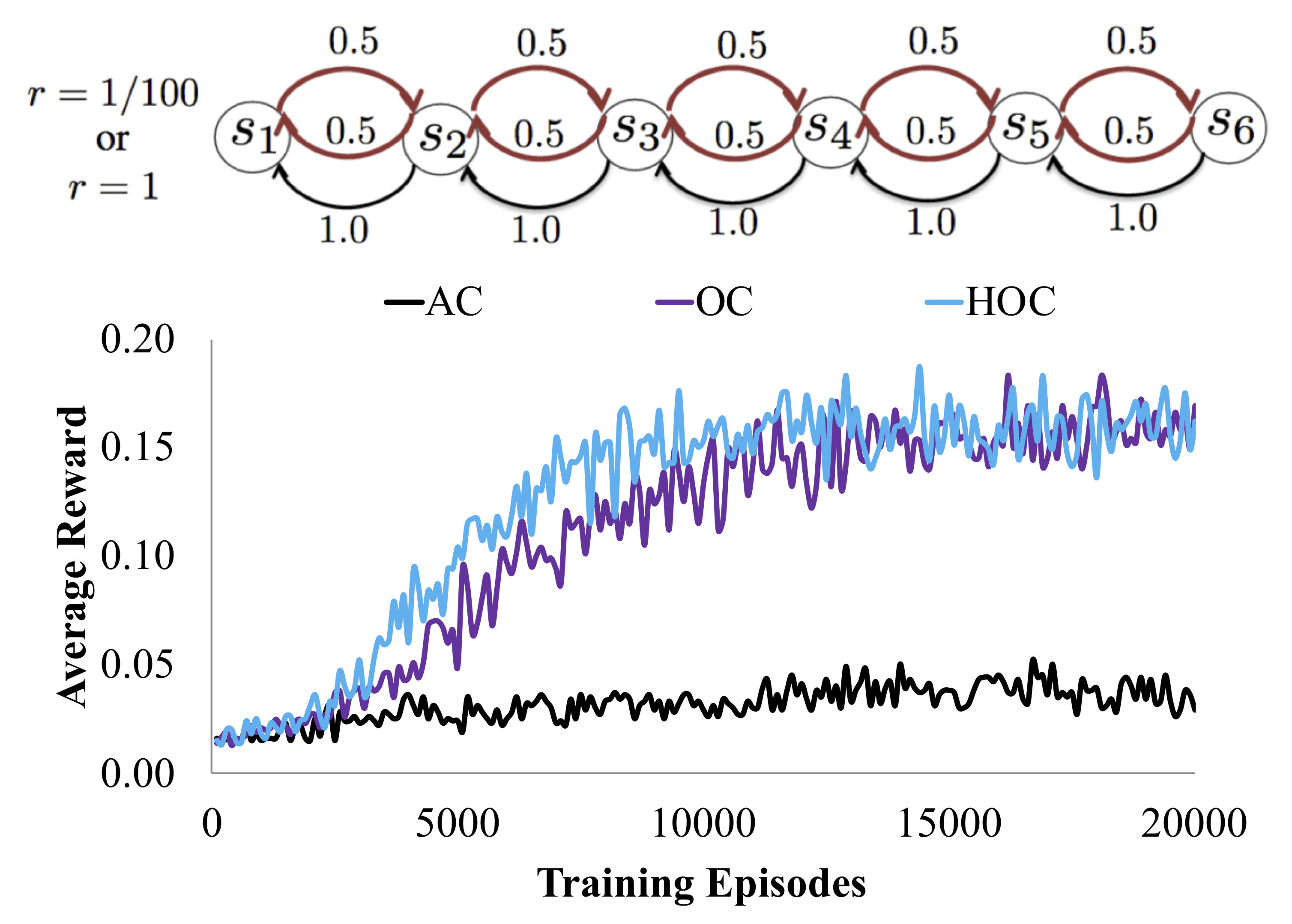}
	\vspace{-6mm}
\caption{The diagram, from \citep{hDQN}, details the stochastic decision process challenge problem. The chart compares learning performance across abstract reasoning levels.}
\vspace{-3mm}
	\label{SMDP}
\end{wrapfigure}

\textbf{Exploring four rooms:} We first consider a navigation task in the four-rooms domain \citep{Options}. Our goal is to evaluate the ability of a set of options learned fully autonomously to learn an efficient exploration policy within the environment. The initial state and the goal state are drawn uniformly from all open non-wall cells every episode. This setting is highly non-stationary, since the goal changes every episode. Primitive movements can fail with probability $\frac{1}{3}$, in which case the agent transitions randomly to one of the empty adjacent cells. The reward is +1 at the goal and 0 otherwise.  In Figure \ref{Fourrooms} we report the average number of steps taken in the last 100 episodes every 100 episodes, reporting the average of 50 runs with different random seeds for each algorithm. We can clearly see that reasoning with higher levels of abstraction is critical to achieving a good exploration policy and that reasoning with three levels of abstraction results in better sample efficient learning than reasoning with two levels of abstraction. For this experiment we explore four levels of abstraction as well, but unfortunately there seem to be diminishing returns at least for this tabular setting.  

\textbf{Discrete stochastic decision process:} Next, we consider a hierarchical RL challenge problem as explored in \citep{hDQN} with a stochastic decision  process  where  the reward depends on the history of visited states in addition to the current state. There are 6 possible states and the agent always starts at $s_2$.  The agent moves left deterministically when it chooses left action; but the action right only succeeds half of the time, resulting in a left move otherwise.  The terminal state is $s_1$ and the agent receives a reward of 1 when it first visits $s_6$ and then $s_1$. The reward for going to $s_1$ without visiting $s_6$ is 0.01. In Figure \ref{SMDP} we report the average reward over the last 100 episodes every 100 episodes, considering 10 runs with different random seeds for each algorithm. Reasoning with higher levels of abstraction is again critical to performing well at this task with reasonable sample efficiency. Both OC learning and HOC learning converge to a high quality solution surpassing performance obtained in \citep{hDQN}. However, it seems that learning converges faster with three levels of abstractions than it does with just two. 

\vspace{-2mm}
\subsection{Deep Function Approximation Problems}

\begin{wrapfigure}{r}{0.45\textwidth}
\vspace{-5mm}
	\includegraphics[scale=0.21]{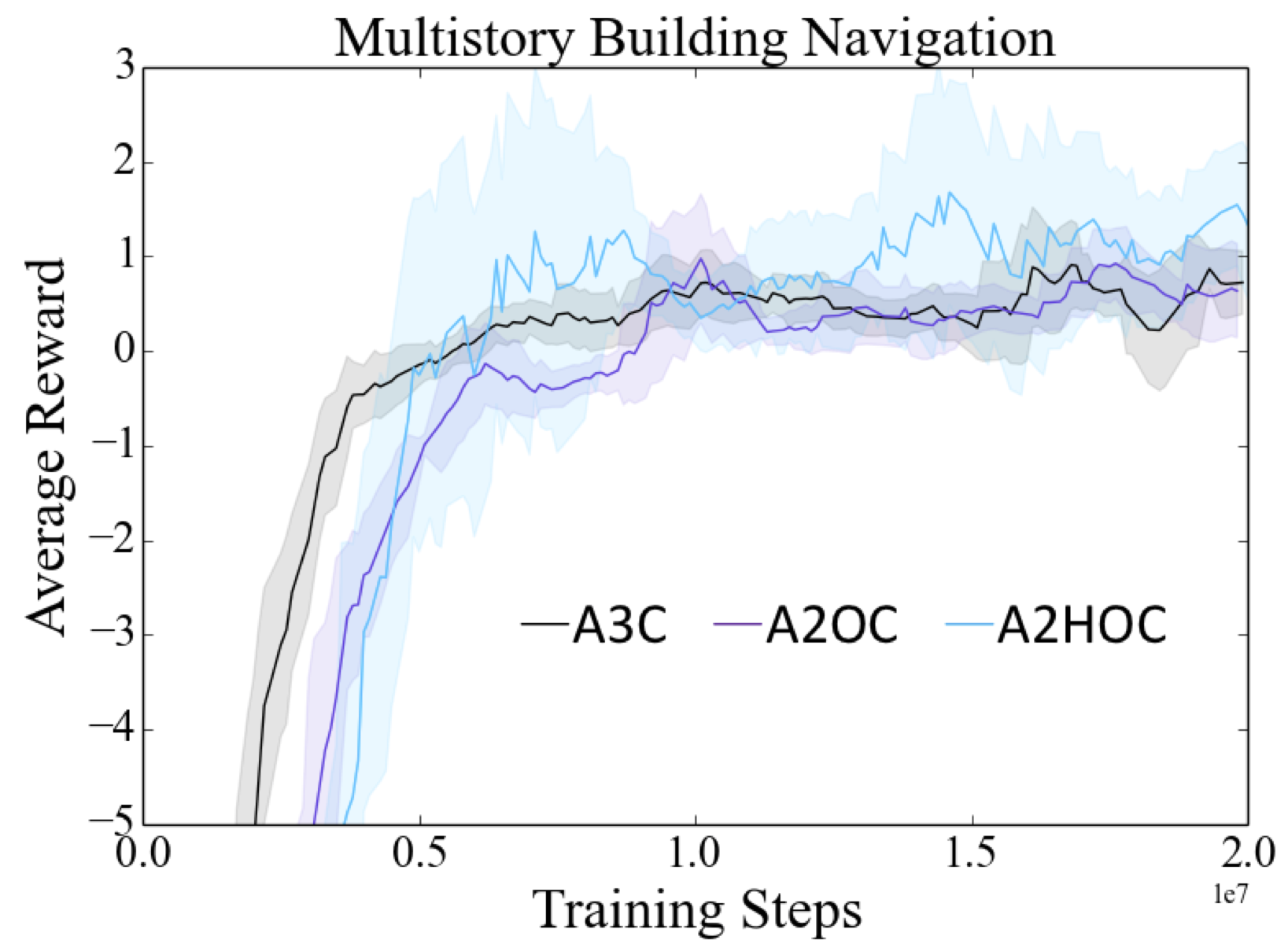}
	\vspace{-3mm}
\caption{Building navigation learning performance across abstract reasoning levels.}
\vspace{-3mm}
	\label{POMDP}
\end{wrapfigure}

\textbf{Multistory building navigation:} For an intuitive look at higher level reasoning, we consider the four rooms problem in a partially observed setting with an 11x17 grid at each level of a seven level building. The agent has a receptive field size of 3 in both directions, so observations for the agent are 9-dimension feature vectors with 0 in empty spots, 1 where there is a wall, 0.25 if there are stairs, or 0.5 if there is a goal location. The stairwells in the north east corner of the floor lead upstairs to the south west corner of the next floor up. Stairs in the south west corner of the floor lead down to the north east corner of the floor below. Agents start in a random location in the basement (which has no south west stairwell) and must navigate to the roof (which has no north east stairwell) to find the goal in a random location. The reward is +10 for finding the goal and -0.1 for hitting a wall. This task could seemingly benefit from abstraction such as a composition of sub-policies to get to the stairs at each intermediate level. We report the rolling mean and standard deviation of the reward. In Figure \ref{POMDP} we see a qualitative difference between the policies learned with three levels of abstraction which has high variance, but fairly often finds the goal location and those learned with less abstraction. A2OC and A3C are hovering around zero reward, which is equivalent to just learning a policy that does not run into walls.  

\begin{wraptable}{r}{5.5cm}
\vspace{-4mm}
\centering
\resizebox{0.4\columnwidth}{!}{
\begin{tabular}{l|c}
\toprule
Architecture & Clipped Reward  \\ \hline
A3C & 8.43 \tiny{$\pm$ 2.29} \\ 
A2OC & 10.56 \tiny{$\pm$0.49} \\ 
A2HOC & \textbf{13.12}  \tiny{$\pm$1.46} \\
\bottomrule
\end{tabular}}
\caption{Average clipped reward per episode over 5 runs on 21 Atari games.} \label{MTE}
\vspace{-2mm}
\end{wraptable}

\textbf{Learning many Atari games with one model:} We finally consider application of the HOC to the Atari games \citep{ALE}. Evaluation protocols for the Atari games are famously inconsistent \citep{AtariEval}, so to ensure for fair comparisons we implement apples to apples versions of our baseline architectures deployed with the same code-base and environment settings. We put our models to the test and consider a very challenging setting \citep{MTL} where a single agent attempts to learn many Atari games at the same time. Our agents attempt to learn 21 Atari games simultaneously, matching the largest previous multi-task setting on Atari \citep{MTL}. Our tasks are hand picked to fall into three categories of related games each with 7 games represented. The first category is games that include maze style navigation (e.g. MsPacman), the second category is mostly fully observable shooter games (e.g. SpaceInvaders), and the final category is partially observable shooter games (e.g. BattleZone). We train each agent by always sampling the game with the least training frames after each episode, ensuring the games are sampled very evenly throughout training. We also clip rewards to allow for equal learning rates across tasks \citep{DQN}. We train each game for 10 million frames (210 million total) and report statistics on the clipped reward achieved by each agent when evaluating the policy without learning for another 3 million frames on each game across 5 separate training runs. As our main metric, we report the summary of how each multi-task agent maximizes its reward in Table \ref{MTE}. While all agents struggle in this difficult setting, HOC is better able to exploit commonalities across games using fewer parameters and policies. 

\begin{figure}
	\includegraphics[scale=0.2]{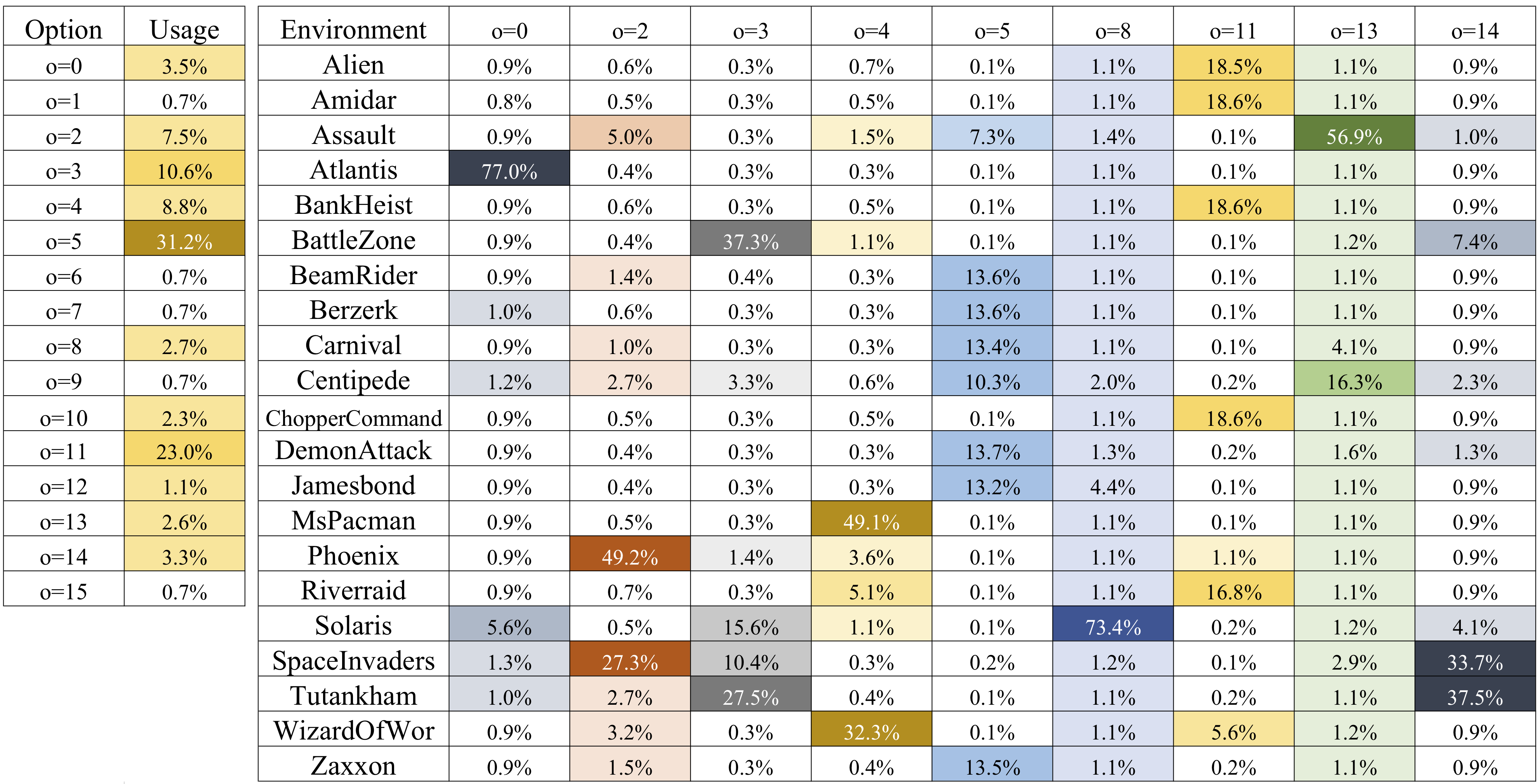}
	\vspace{-7mm}
\caption{Option usage (left) and specialization across Atari games for the top 9 most used options (right) of a 16 option Option-Critic architecture trained in the many task learning setting.}
	\label{OCOptions}
	\vspace{-2mm}
\end{figure}

\begin{figure}
	\includegraphics[scale=0.2]{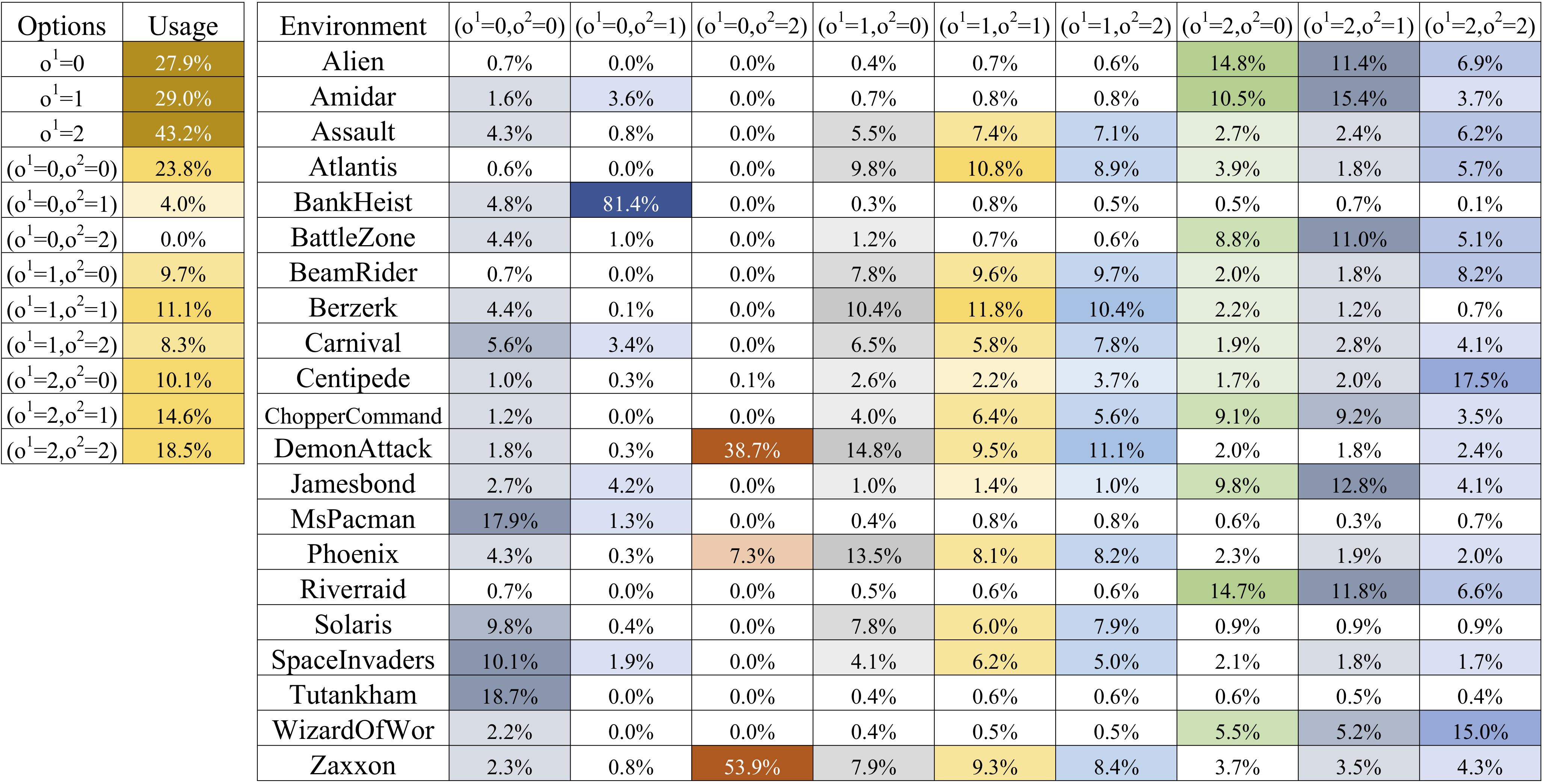} 
	\vspace{-6mm}
\caption{Option usage (left) and specialization across Atari games (right) of a Hierarchical Option-Critic architecture with $N=3$ and 3 options at each layer trained in the many task learning setting.}
	\label{HOCOptions}
	\vspace{-5mm}
\end{figure}

\textbf{Analysis of Learned Options:} An advantage of the multi-task setting is it allows for a degree of quantitative interpretability regarding when and how options are used. We report characteristics of the agents with median performance during the evaluation period. A2OC with 16 options uses 5 options the bulk of the time with the rest of the time largely split among another 6 options (Figure \ref{OCOptions}). The average number of time steps between switching options has a pretty narrow range across games falling between 3.4 (Solaris) and 5.5 (MsPacman). In contrast, A2HOC with three options at each branch of the hierarchy learns to switch options at a rich range of temporal resolutions depending on the game. The high level options vary between an average of 3.2 (BeamRider) and 9.7 (Tutankham) steps before switching. Meanwhile, the low level options vary between an average of 1.5 (Alien) and 7.8 (Tutankham) steps before switching. In Appendix \ref{MTL} we provide additional details about the average duration before switching options for each game. In Figure \ref{HOCOptions} we can see that the most used options for HOC are distributed pretty evenly across a number of games, while OC tends to specialize its options on a smaller number of games. In fact, the average share of usage dominated by a single game for the top 7 most used options is 40.9\% for OC and only 14.7\% for HOC. Additionally, we can see that a hierarchy of options imposes structure in the space of options. For example, when $o^1=1$ or $o^1=2$ the low level options tend to focus on different situations within the same games. 

\vspace{-1mm}
\section{Conclusion}
\vspace{-1mm}
In this work we propose the first policy gradient theorems to optimize an arbitrarily deep hierarchy of options to maximize the expected discounted return. Moreover, we have proposed a particular hierarchical option-critic architecture that is the first general purpose reinforcement learning architecture to successfully learn options from data with more than two abstraction levels. We have conducted extensive empirical evaluation in the tabular and deep non-linear function approximation settings. In all cases we found that, for significantly complex problems, reasoning with more than two levels of abstraction can be beneficial for learning. While the performance of the hierarchical option-critic architecture is impressive, we envision our proposed policy gradient theorems eventually transcending it in overall impact. Although the architectures we explore in this paper have a fixed structure and fixed depth of abstraction for simplicity, the underlying theorems can also guide learning for much more dynamic architectures that we hope to explore in future work. 

\section*{Acknowledgements}
	The authors thank Murray Campbell, Xiaoxiao Guo, Ignacio Cases, and Tim Klinger for fruitful discussions that helped shape this work. We also would like to thank Modjtaba Shokrian-Zini and Ravi Chunduru for their comments after publication that helped shape our revised draft. 
\bibliographystyle{plainnat}
\bibliography{nips_2017}

\appendix

\section{Derivation of Generalized Policy Gradient and Termination Gradient Theorems} \label{Proof}

\subsection{The Derivation of U}  \label{Proof:0}

To help explain the meaning and derivation of equation \eqref{Ueq}, we separate the expression into four primary terms. The first term is applicable for $N \geq 1$ and represents the expected return from cases where no options terminate. The second term is applicable for $N \geq 2$ and represents the expected return from cases where every option terminates. The third and fourth terms are applicable for $N \geq 3$ and represent the expected return from cases where some options terminate. 

We will first discuss how to estimate the return when there are no terminated options. In this case we simply use our estimate of the value of the current state following the current options if there are any. As we are computing the expectation, we also multiply this term by its likelihood of happening which is equal to the probability that the lowest level option policy does not terminate. When $N=1$ we can consider the termination probability of the current policy as zero and the current option context to be empty. As such, we estimate the value function upon arrival as $V_\Omega(s)$ as we do for actor-critic policy gradients. 

Next we turn our attention to estimating the return when all options are terminated. This can be approximated using our estimate of the return given the state $V_\Omega(s)$. The likelihood of this happening is equal to the conditional likelihood of options terminating at every level of abstraction we are modeling. When $N=2$, equation \eqref{Ueq} simplifies to equation \eqref{U2}. This expression is precisely the option value function upon arrival of the option-critic framework derived in \citep{OC}. 

The final quantity we will estimate bridges the gap to cases where only some options terminate. This situation has not been explored by other work on option learning as it only arises for situations with at least $N=3$ hierarchical levels of planning. The case where some (but not all) options terminate arises when a series of low level options terminate while a high level option does not terminate. For a given level of abstraction, we can analyze the likelihood that at each level the lower level options terminate while the current does not. In such a case, we multiply this likelihood by the value one level more abstract than the current option hierarchy level. For convenience in our derivation, we split our notation for this quantity into two separate terms accounting explicitly for the case when only lower level options terminate.  

\subsection{Generalized Markov Chain and Augmented Process}  \label{Proof:1}

We must establish the Markov chain along which we can measure performance for options with $N$ levels of abstraction. The natural  approach  is  to  consider  the  chain  defined  in  the  augmented state space because state and active option based tuples now play the role of regular states in a usual Markov chain. If options $o_t^{1:N-1}$ have been initiated or are executing at time $t$ in state $s_t$, then the probability of transitioning to $(s_{t+1},o_{t+1}^{1:\ell-1})$ in one step is:

\vspace{-5mm}
\begin{equation} \label{PMC}
\begin{split}
P(s_{t+1},o_{t+1}^{1:\ell-1}|s_t,o_t^{1:N-1}) = \sum_{o_t^N} \pi_{\theta^N}^N(o_t^N|s_t,o_t^{1:N-1}) P(s_{t+1}|s_t,o_t^{N}) [ \\ \underbrace{(1-\beta_{\phi^{N-1}}^{N-1}(s_{t+1},o_t^{1:N-1})) \textbf{1}_{o_{t+1}^{1:\ell-1}=o_{t}^{1:\ell-1}} }_{\text{none terminate}} + \\ \underbrace{ \sum_{q=N-1}^\ell (1-\beta_{\phi^{q-1}}^{q-1}(s_{t+1},o_t^{1:q-1})) \prod_{z=N-1}^q \beta_{\phi^z}^z(s_{t+1},o_t^{1:z}) \textbf{1}_{o_{t+1}^{1:\ell-1}=o_{t}^{1:\ell-1}} }_{\text{only lower level options terminate}} + \\  \underbrace{\prod_{j=N-1}^{1} \beta_{\phi^j}^j(s_{t+1},o_t^{1:j}) \prod_{v=\ell-1}^{1} \pi_{\theta^v}^v(o_{t+1}^v|s_{t+1},o_t^{1:v-1})}_{\text{all options terminate}}  +  \\
\underbrace{ \sum_{i=1}^{\ell - 2}(1-\beta_{\phi^i}^i(s_{t+1},o_t^{1:i})) \prod_{k=i+1}^{N-1} \beta_{\phi^k}^k(s_{t+1},o_t^{1:k}) \prod_{p=i+1}^{\ell-1} \pi_{\theta^p}^p(o_{t+1}^p|s_{t+1},o_t^{1:p-1})}_{\text{some relevant higher level options terminate}}].
\end{split}
\end{equation}

where primitive actions are $o^N$. Like the Markov chain derived for the option critic architecture \citep{OC}, the process given by equation \eqref{PMC} is homogeneous. Additionally, when options are available at every state, the process is ergodic with the existance of a unique stationary distribution over the augmented state space tuples.   

We continue by presenting an extension of results about augmented processes used for derivation of learning algorithms in \citep{OC} to an option hierarchy with $N$ levels of abstraction. If options $o^{1:N-1}_t$ have been initiated or are executing at time $t$, then the discounted probability of transitioning to $(s_{t+1},o^{1:\ell-1}_{t+1})$ where $\ell \leq N$ is:

\vspace{-5mm}
\begin{equation} \label{DMC}
\begin{split}
P_\gamma^{(1)}(s_{t+1},o_{t+1}^{1:\ell-1}|s_t,o_t^{1:N-1}) = \sum_{o_t^N} \pi_{\theta^N}^N(o_t^N|s_t,o_t^{1:N-1})  \gamma P(s_{t+1}|s_t,o_t^{N}) [ \\ \underbrace{(1-\beta_{\phi^{N-1}}^{N-1}(s_{t+1},o_t^{1:N-1})) \textbf{1}_{o_{t+1}^{1:\ell-1}=o_{t}^{1:\ell-1}} }_{\text{none terminate}} + \\ \underbrace{ \sum_{q=N-1}^\ell (1-\beta_{\phi^{q-1}}^{q-1}(s_{t+1},o_t^{1:q-1})) \prod_{z=N-1}^q \beta_{\phi^z}^z(s_{t+1},o_t^{1:z}) \textbf{1}_{o_{t+1}^{1:\ell-1}=o_{t}^{1:\ell-1}} }_{\text{only lower level options terminate}} + \\  \underbrace{\prod_{j=N-1}^{1} \beta_{\phi^j}^j(s_{t+1},o_t^{1:j}) \prod_{v=\ell-1}^{1} \pi_{\theta^v}^v(o_{t+1}^v|s_{t+1},o_t^{1:v-1})}_{\text{all options terminate}}  +  \\
\underbrace{ \sum_{i=1}^{\ell - 2}(1-\beta_{\phi^i}^i(s_{t+1},o_t^{1:i})) \prod_{k=i+1}^{N-1} \beta_{\phi^k}^k(s_{t+1},o_t^{1:k}) \prod_{p=i+1}^{\ell-1} \pi_{\theta^p}^p(o_{t+1}^p|s_{t+1},o_t^{1:p-1})}_{\text{some relevant higher level options terminate}}]. \\
\end{split}
\end{equation}
As such, when we  condition  the  process  from $(s_{t},o^{1:N-1}_{t-1})$,  the  discounted probability of transitioning to $(s_{t+1},o^{1:\ell-1}_{t})$ is:

\begin{equation}  \label{P_PG}
\begin{split}
P_\gamma^{(1)}(s_{t+1},o_{t}^{1:\ell-1}|s_t,o_{t-1}^{1:N-1}) = \sum_{o_t^N} \pi_{\theta^N}^N(o_t^N|s_t,o_t^{1:N-1})  \gamma P(s_{t+1}|s_t,o_t^{N}) [ \\ \underbrace{(1-\beta_{\phi^{N-1}}^{N-1}(s_{t+1},o_{t-1}^{1:N-1})) \textbf{1}_{o_t^{1:\ell-1}=o_{t-1}^{1:\ell-1}} }_{\text{none terminate}} + \\ \underbrace{ \sum_{q=N-1}^\ell (1-\beta_{\phi^{q-1}}^{q-1}(s_{t+1},o_{t-1}^{1:q-1})) \prod_{z=N-1}^q \beta_{\phi^z}^z(s_{t+1},o_{1:t-1}^{z}) \textbf{1}_{o_t^{1:\ell-1}=o_{t-1}^{1:\ell-1}} }_{\text{only lower level options terminate}} + \\  \underbrace{\prod_{j=N-1}^{1} \beta_{\phi^j}^j(s_{t+1},o_{t-1}^{1:j}) \prod_{v=\ell-1}^{1} \pi_{\theta^v}^v(o_{t}^v|s_{t+1},o_{t-1}^{1:v-1})}_{\text{all options terminate}}  +  \\
\underbrace{ \sum_{i=1}^{\ell - 2}(1-\beta_{\phi^i}^i(s_{t+1},o_{t-1}^{1:i})) \prod_{k=i+1}^{N-1} \beta_{\phi^k}^k(s_{t+1},o_{t-1}^{1:k}) \prod_{p=i+1}^{\ell-1} \pi_{\theta^p}^p(o_{t}^p|s_{t+1},o_{t-1}^{1:p-1})}_{\text{some relevant higher level options terminate}}]. \\
\end{split}
\end{equation}

This definition will be very useful later for our derivation of the hierarchical intra-option policy gradient. However, for the derivation of the hierarchical termination gradient theorem we should reformulate the discounted probability of transitioning to $(s_{t+1},o^{1:\ell}_{t})$ from the view of the termination policy at abstraction level $\ell$ explicitly separating out terms that depend on $\phi^\ell$:

\begin{equation}  \label{T_PG}
\begin{split}
P_\gamma^{(1)}(s_{t+1},o_{t}^{1:\ell}|s_t,o_{t-1}^{1:N-1}) = \sum_{o_t^N} \pi_{\theta^N}^N(o_t^N|s_t,o_t^{1:N-1})  \gamma P(s_{t+1}|s_t,o_t^{N}) [ \\
\underbrace{(1-\beta_{\phi^{N-1}}^{N-1}(s_{t+1},o_{t-1}^{1:N-1})) \textbf{1}_{o_t^{1:\ell}=o_{t-1}^{1:\ell}} }_{\text{none terminate}} + \underbrace{ \sum_{q=N-1}^{\ell+2} (1-\beta_{\phi^{q-1}}^{q-1}(s_{t+1},o_{t-1}^{1:q-1})) \prod_{z=N-1}^q \beta_{\phi^z}^z(s_{t+1},o_{t-1}^{1:z}) \textbf{1}_{o_t^{1:\ell}=o_{t-1}^{1:\ell}} }_{\text{only lower level options terminate}} + \\ \underbrace{ (1-\beta_{\phi^{\ell}}^{\ell}(s_{t+1},o_{t-1}^{1:\ell})) \prod_{z=N-1}^{\ell+1} \beta_{\phi^z}^z(s_{t+1},o_{t-1}^{1:z}) \textbf{1}_{o_t^{1:\ell}=o_{t-1}^{1:\ell}} }_{\text{$\ell+1$ terminates and $\ell$ does not}} +  \underbrace{\prod_{j=N-1}^{1} \beta_{\phi^j}^j(s_{t+1},o_{t-1}^{1:j}) \prod_{v=\ell}^{1} \pi_{\theta^v}^v(o_{t}^v|s_{t+1},o_{t-1}^{1:v-1})}_{\text{all options terminate}}  +  \\
\underbrace{ \sum_{i=1}^{\ell - 1}(1-\beta_{\phi^i}^i(s_{t+1},o_{t-1}^{1:i})) \prod_{k=i+1}^{N-1} \beta_{\phi^k}^k(s_{t+1},o_{t-1}^{1:k}) \prod_{p=i+1}^{\ell} \pi_{\theta^p}^p(o_{t}^p|s_{t+1},o_{t-1}^{1:p-1})}_{\text{some relevant higher level options terminate}}]. \\
\end{split}
\end{equation}

The $k$-step discounted probabilities can more generally be expressed recursively:

\vspace{-4mm}
\begin{equation}
\begin{split}
P_\gamma^{(k)}(s_{t+k},o_{t+k}^{1:\ell-1}|s_{t},o_{t}^{1:N-1}) = \\ \sum_{s_{t+1}} \sum_{o_{t+1}^1} ... \sum_{o_{t+1}^{N-1}} [  P_\gamma^{(1)}(s_{t+1},o_{t+1}^{1:N-1}|s_{t},o_{t}^{1:N-1}) P_\gamma^{(k-1)}(s_{t+k-1},o_{t+k}^{1:\ell-1}|s_{t+1},o_{t+1}^{1:N-1}) ].
\end{split}
\end{equation}
\vspace{-4mm}

Or rather conditioning on $t-1$ as in equation \eqref{P_PG}:

\vspace{-4mm}
\begin{equation}
\begin{split}
P_\gamma^{(k)}(s_{t+k},o_{t+k-1}^{1:\ell-1}|s_{t},o_{t-1}^{1:N-1}) =  \\ \sum_{s_{t+1}} \sum_{o_{t}^1} ... \sum_{o_{t}^{1:N-1}} [ P_\gamma^{(1)}(s_{t+1},o_{t}^{1:N-1}|s_{t},o_{t-1}^{1:N-1})  P_\gamma^{(k-1)}(s_{t+k-1},o_{t+k-1}^{1:\ell-1}|s_{t+1},o_{t}^{1:N-1}) ].
\end{split}
\end{equation}
\vspace{-4mm}

\subsection{Proof of the Hierarchical Intra-Option Policy Gradient Theorem}  \label{Proof:2}

Taking the gradient of the value function with an augmented state space:

\vspace{-5mm}
\begin{equation} 
\begin{split}
\frac{\partial Q_\Omega(s,o^{1:\ell-1})}{\partial \theta^\ell} = \frac{\partial}{\partial \theta^\ell} \sum_{o^\ell} \pi^\ell_{\theta^\ell}(o^\ell|s,o^{1:\ell-1})Q_U(s,o^{1:\ell}) \\
= \sum_{o^\ell} (\frac{\partial\pi^\ell_{\theta^\ell}(o^\ell|s,o^{1:\ell-1})}{\partial \theta^\ell}Q_U(s,o^{1:\ell}) + \pi^\ell_{\theta^\ell}(o^\ell|s,o^{1:\ell-1})\frac{\partial Q_U(s,o^{1:\ell})}{\partial \theta^\ell})  
\end{split}
\end{equation}
\vspace{-5mm}

Then substituting in equation \ref{QwithU} with the assumption that $\theta^\ell$ only appears in the intra-option policy at level $\ell$ and not in any policy at another level or in the termination function:

\begin{equation} \label{PG1}
\begin{split}
\frac{\partial Q_\Omega(s,o^{1:\ell-1})}{\partial \theta^\ell} = \sum_{o^\ell} (\frac{\partial\pi^\ell_{\theta^\ell}(o^\ell|s,o^{1:\ell-1})}{\partial \theta^\ell}Q_U(s,o^{1:\ell}) +
\pi^\ell_{\theta^\ell}(o^\ell|s,o^{1:\ell-1}) \gamma \sum_{s'} P(s'|s,o^{1:\ell}) \frac{\partial U(s',o^{1:\ell-1})}{\partial \theta^\ell})
\end{split}
\end{equation}

where $P(s'|s,o^{1:\ell})$ is the probability of transitioning to a state based on the augmented state space $(s,o^{1:\ell})$ considering primitive actions $o^N$:

\vspace{-5mm}
\begin{equation} \label{simplify}
P(s'|s,o^{1:\ell}) = \sum_{o^N}...\sum_{o^{\ell+1}} P(s'|s,o^N) \prod_{j=\ell+1}^N \pi^j(o^j|s,o^{1:j-1}).
\end{equation}
\vspace{-3mm}

We continue by computing the gradient with respect to $U$ again assuming that $\theta^\ell$ only appears in the intra-option policy at level $\ell$ and not in any policy at another level or in the termination function:

\begin{equation} 
\begin{split}
\frac{\partial U(s',o^{1:\ell-1})}{\partial \theta^\ell} = \underbrace{(1-\beta_{\phi^{N-1}}^{N-1}(s',o^{1:N-1})) \frac{\partial Q_\Omega(s',o^{1:\ell-1})}{\partial \theta^\ell}}_{\text{none terminate}} + \underbrace{ \frac{\partial V_\Omega(s')}{\partial \theta^\ell} \prod_{j=N-1}^{1} \beta_{\phi^j}^j(s',o^{1:j})}_{\text{all options terminate}}  + \\
\underbrace{ \frac{\partial Q_\Omega(s',o^{1:\ell-1})}{\partial \theta^\ell} \sum_{q=N-1}^\ell (1-\beta_{\phi^{q-1}}^{q-1}(s',o^{1:q-1})) \prod_{z=N}^q \beta_{\phi^z}^z(s',o^{1:z})}_{\text{only lower level options terminate}} + \\
\underbrace{ \sum_{i=1}^{\ell - 2}(1-\beta_{\phi^i}^i(s',o^{1:i})) \frac{\partial Q_\Omega(s',o^{1:i})}{\partial \theta^\ell}\prod_{k=i+1}^{N-1} \beta_{\phi^k}^k(s',o^{1:k}) }_{\text{some relevant higher level options terminate}} \\
\end{split}
\end{equation}

Next we integrate out the lower level options so that each term is operating in the same augmented state space: 
\begin{equation} 
\begin{split}
\frac{\partial U(s',o^{1:\ell-1})}{\partial \theta^\ell} = \underbrace{(1-\beta_{\phi^{N-1}}^{N-1}(s',o^{1:N-1})) \frac{\partial Q_\Omega(s',o^{1:\ell-1})}{\partial \theta^\ell} }_{\text{none terminate}} + \\ \underbrace{ \sum_{q=N-1}^\ell (1-\beta_{\phi^{q-1}}^{q-1}(s',o^{1:q-1})) \prod_{z=N-1}^q \beta_{\phi^z}^z(s',o^{1:z}) \frac{\partial Q_\Omega(s',o^{1:\ell-1})}{\partial \theta^\ell} }_{\text{only lower level options terminate}} + \\ \underbrace{ \prod_{j=N-1}^{1} \beta_{\phi^j}^j(s',o^{1:j}) \sum_{o'^{1}} ... \sum_{o'^{\ell-1}} \prod_{v=\ell-1}^{1} \pi_{\theta^v}^v(o'^v|s',o'^{1:v-1}) \frac{ \partial Q_\Omega(s',o'^{1:\ell-1})}{\partial \theta^\ell}}_{\text{all options terminate}}  +  \\ 
\underbrace{ \sum_{i=1}^{\ell - 2}(1-\beta_{\phi^i}^i(s',o^{1:i})) \prod_{k=i+1}^{N-1} \beta_{\phi^k}^k(s',o^{1:k}) \sum_{o'^{i+1}} ... \sum_{o'^{\ell-1}} \prod_{p=i+1}^{\ell-1} \pi_{\theta^p}^p(o'^p|s',o'^{1:p-1}) \frac{ \partial Q_\Omega(s',o'^{1:\ell-1})}{\partial \theta^\ell}}_{\text{some relevant higher level options terminate}} 
\end{split}
\end{equation}

We can then simplify our expression: 

\begin{equation} \label{PG2}
\begin{split}
\frac{\partial U(s',o^{1:\ell-1})}{\partial \theta^\ell}  = \sum_{o'^1}...\sum_{o'^\ell-1} [ \underbrace{(1-\beta_{\phi^{N-1}}^{N-1}(s',o^{1:N-1})) \textbf{1}_{o'^{1:\ell-1}=o^{1:\ell-1}} }_{\text{none terminate}} + \\ \underbrace{ \sum_{q=N-1}^\ell (1-\beta_{\phi^{q-1}}^{q-1}(s',o^{1:q-1})) \prod_{z=N}^q \beta_{\phi^z}^z(s',o^{1:z}) \textbf{1}_{o'^{1:\ell-1}=o^{1:\ell-1}} }_{\text{only lower level options terminate}} + \\ \underbrace{\prod_{j=N-1}^{1} \beta_{\phi^j}^j(s',o^{1:j}) \prod_{v=\ell-1}^{1} \pi_{\theta'^v}^v(o'^v|s',o'^{1:v-1})}_{\text{all options terminate}}  +  \\ 
\underbrace{ \sum_{i=1}^{\ell - 2}(1-\beta_{\phi^i}^i(s',o^{1:i})) \prod_{k=i+1}^{N-1} \beta_{\phi^k}^k(s',o^{1:k}) \prod_{p=i+1}^{\ell-1} \pi_{\theta^p}^p(o'^p|s',o'^{1:p-1})}_{\text{some relevant higher level options terminate}}] \frac{ \partial Q_\Omega(s',o'^{1:\ell-1})}{\partial \theta^\ell}, 
\end{split}
\end{equation}
We proceed by substituting \eqref{PG2} into \eqref{PG1}:  

\begin{equation}
\begin{split}
\frac{\partial Q_\Omega(s,o^{1:\ell-1})}{\partial \theta^\ell} = \sum_{o^\ell} (\frac{\partial\pi^\ell_{\theta^\ell}(o^\ell|s,o^{1:\ell-1})}{\partial \theta^\ell}Q_U(s,o^{1:\ell}) + \\
\pi^\ell_{\theta^\ell}(o^\ell|s,o^{1:\ell-1}) \gamma \sum_{s'} P(s'|s,o^{1:\ell}) \sum_{o'^1}...\sum_{o'^{\ell-1}} [ \underbrace{(1-\beta_{\phi^{N-1}}^{N-1}(s',o^{1:N-1})) \textbf{1}_{o'^{1:\ell-1}=o^{1:\ell-1}} }_{\text{none terminate}} + \\ \underbrace{ \sum_{q=N-1}^\ell (1-\beta_{\phi^{q-1}}^{q-1}(s',o^{1:q-1})) \prod_{z=N}^q \beta_{\phi^z}^z(s',o^{1:z}) \textbf{1}_{o'^{1:\ell-1}=o^{1:\ell-1}} }_{\text{only lower level options terminate}} + \\ \underbrace{\prod_{j=N-1}^{1} \beta_{\phi^j}^j(s',o^{1:j}) \prod_{v=\ell-1}^{1} \pi_{\theta^v}^v(o'^v|s',o'^{1:v-1})}_{\text{all options terminate}}  +  \\ 
\underbrace{ \sum_{i=1}^{\ell - 2}(1-\beta_{\phi^i}^i(s',o^{1:i})) \prod_{k=i+1}^{N-1} \beta_{\phi^k}^k(s',o^{1:k}) \prod_{p=i+1}^{\ell-1} \pi_{\theta^p}^p(o'^p|s',o'^{1:p-1})}_{\text{some relevant higher level options terminate}}] \frac{ \partial Q_\Omega(s',o'^{1:\ell-1})}{\partial \theta^\ell} \\
\end{split}
\end{equation}

This yields a recursion, which can be further simplified to: 

\begin{equation}
\begin{split}
\frac{\partial Q_\Omega(s,o^{1:\ell-1})}{\partial \theta^\ell} &= \\ \sum_{o^\ell}& \frac{\partial\pi^\ell_{\theta^\ell}(o^\ell|s,o^{1:\ell-1})}{\partial \theta^\ell}Q_U(s,o^{1:\ell}) + 
\sum_{s'} \sum_{o'^1} ... \sum_{o'^{\ell-1}} P_\gamma^{(1)}(s',o'^{1:\ell-1}|s,o^{1:N-1}) \frac{ \partial Q_\Omega(s',o'^{1:\ell-1})}{\partial \theta^\ell} 
\end{split}
\end{equation}

Considering the previous remarks about augmented processes and substituting in equation \eqref{P_PG}, this expression becomes:

\begin{equation}
\begin{split}
\frac{\partial Q_\Omega(s,o^{1:\ell-1})}{\partial \theta^\ell} = \sum_{k=0}^\infty \sum_{s',o'^{1:\ell-1}} P_\gamma^{(k)}(s',o'^{1:\ell-1}|s,o^{1:N-1}) \sum_{o^\ell} \frac{\partial \pi^\ell_{\theta^\ell}(o'^\ell|s',o'^{1:\ell-1})}{\partial \theta^\ell}Q_U(s',o'^{1:\ell})
\end{split}
\end{equation}

The gradient of the expected discounted return with respect to $\theta^\ell$ is then:

\begin{equation}
\begin{split}
\frac{\partial Q_\Omega(s_0,o_0^{1:\ell-1})}{\partial \theta^\ell} =  \sum_{s,o^{1:\ell-1}} \sum_{k=0}^\infty  P_\gamma^{(k)}(s,o^{1:\ell-1}|s_0,o_0^{1:N-1}) \sum_{o^\ell} \frac{\partial \pi^\ell_{\theta^\ell}(o^\ell|s,o^{1:\ell-1})}{\partial \theta^\ell}Q_U(s,o^{1:\ell}) \\
 = \sum_{s,o^{1:\ell-1}}  \mu_\Omega(s,o^{1:\ell-1}|s_0,o_0^{1:N-1}) \sum_{o^\ell} \frac{\partial \pi^\ell_{\theta^\ell}(o^\ell|s,o^{1:\ell-1})}{\partial \theta^\ell}Q_U(s,o^{1:\ell}) .
\end{split}
\end{equation}

\subsection{Proof of the Hierarchical Termination Gradient Theorem}  \label{Proof:3}

The  expected  sum  of  discounted  rewards  originating from augmented state $(s_1,o_0^{1:N-1})$ is defined as:

\begin{equation}
U(s_1,o_0^{1:N-1}) = \mathbb{E}[\sum_{t=1}^\infty \gamma^{t-1} r_t | s_1,o_0^{1:N-1}]
\end{equation}

We start by reformulating $U$ from equation \eqref{Ueq} at level of abstraction $\ell$ rather than $\ell-1$ as follows:

\begin{equation} 
\begin{split}
U(s',o^{1:\ell}) = \underbrace{(1-\beta_{\phi^{N-1}}^{N-1}(s',o^{1:N-1}))Q_\Omega(s',o^{1:\ell})}_{\text{none terminate}} + \underbrace{V_\Omega(s) \prod_{j=N-1}^{1} \beta_{\phi^j}^j(s',o^{1:j})}_{\text{all options terminate}}  + \\
\underbrace{ Q_\Omega(s',o^{1:\ell}) \sum_{q=N-1}^{\ell+1} (1-\beta_{\phi^{q-1}}^{q-1}(s',o^{1:q-1})) \prod_{z=N-1}^q \beta_{\phi^z}^z(s',o^{1:z})}_{\text{only lower level options terminate}} + \\
\underbrace{ \sum_{i=1}^{\ell - 1}(1-\beta_{\phi^i}^i(s',o^{1:i})) Q_\Omega(s',o^{1:i}) \prod_{k=i+1}^{N-1} \beta_{\phi^k}^k(s',o^{1:k}) }_{\text{some relevant higher level options terminate}} \\
\end{split}
\end{equation}

As we will be interested in analyzing this expression with respect to $\phi^\ell$, we separate the term where only lower level options terminate into two separate terms. In the special case where $\ell+1$ terminates and $\ell$ does not, we still utilize $\phi^\ell$ even though it did not terminate:

\begin{equation} 
\begin{split}
U(s',o^{1:\ell}) = \underbrace{(1-\beta_{\phi^{N-1}}^{N-1}(s',o^{1:N-1}))Q_\Omega(s',o^{1:\ell})}_{\text{none terminate}} + 
\underbrace{V_\Omega(s) \prod_{j=N-1}^{1} \beta_{\phi^j}^j(s',o^{1:j})}_{\text{all options terminate}}  + \\
\underbrace{ Q_\Omega(s',o^{1:\ell}) \sum_{q=N-1}^{\ell+2} (1-\beta_{\phi^{q-1}}^{q-1}(s',o^{1:q-1})) \prod_{z=N-1}^q \beta_{\phi^z}^z(s',o^{1:z})}_{\text{only lower level options than $\ell+1$ terminate}} + \\
\underbrace{ Q_\Omega(s',o^{1:\ell}) (1-\beta_{\phi^{\ell}}^{\ell}(s',o^{1:\ell})) \prod_{z=N-1}^{\ell+1} \beta_{\phi^z}^z(s',o^{1:z})}_{\text{$\ell+1$ terminates and $\ell$ does not}} + 
\underbrace{ \sum_{i=1}^{\ell - 1}(1-\beta_{\phi^i}^i(s',o^{1:i})) Q_\Omega(s',o^{1:i}) \prod_{k=i+1}^{N-1} \beta_{\phi^k}^k(s',o^{1:k}) }_{\text{some relevant higher level options terminate}} \\
\end{split}
\end{equation}

The original expression of $U$ was more useful for the gradient with respect to $\theta^\ell$, which does not depend on this case. The gradient of $U$ with respect to $\phi^\ell$ is then: 

\begin{equation} 
\begin{split}
\frac{\partial U(s',o^{1:\ell})}{\partial \phi^\ell} = 
\underbrace{V_\Omega(s) \frac{\partial \beta_{\phi^\ell}^\ell(s',o^{1:\ell})}{\partial \phi^\ell} [\prod_{j=N-1}^{\ell+1} \beta_{\phi^j}^j(s',o^{1:j})] [\prod_{j=\ell-1}^{1} \beta_{\phi^j}^j(s',o^{1:j})]}_{\text{ (1) all options terminate}}  + \\
\underbrace{ Q_\Omega(s',o^{1:\ell}) (-\frac{\partial \beta_{\phi^{\ell}}^{\ell}(s',o^{1:\ell})}{\partial \phi^\ell}) \prod_{z=N-1}^{\ell+1} \beta_{\phi^z}^z(s',o^{1:z})}_{\text{ (2) $\ell+1$ terminates and $\ell$ does not}} + \\
\underbrace{ \sum_{i=1}^{\ell - 1}(1-\beta_{\phi^i}^i(s',o^{1:i})) Q_\Omega(s',o^{1:i}) \frac{\partial \beta_{\phi^\ell}^\ell(s',o^{1:\ell})}{\partial \phi^\ell} [\prod_{k=i+1}^{\ell - 1} \beta_{\phi^k}^k(s',o^{1:k})][\prod_{k=\ell+1}^{N - 1} \beta_{\phi^k}^k(s',o^{1:k})] }_{\text{ (3) some relevant higher level options terminate}} +\\
\underbrace{(1-\beta_{\phi^{N-1}}^{N-1}(s',o^{1:N-1})) \frac{\partial Q_\Omega(s',o^{1:\ell})}{\partial \phi^\ell}}_{\text{ (4) none terminate}} + \\
+ \underbrace{ \frac{\partial V_\Omega(s)}{\partial \phi^\ell} \prod_{j=N-1}^{1} \beta_{\phi^j}^j(s',o^{1:j})}_{\text{ (5) all options terminate}}  + 
\underbrace{ \frac{\partial Q_\Omega(s',o^{1:\ell})}{\partial \phi^\ell} \sum_{q=N-1}^{\ell+2} (1-\beta_{\phi^{q-1}}^{q-1}(s',o^{1:q-1})) \prod_{z=N-1}^q \beta_{\phi^z}^z(s',o^{1:z})}_{\text{ (6) only lower level options than $\ell+1$ terminate}} + \\
\underbrace{ \frac{\partial Q_\Omega(s',o^{1:\ell})}{\partial \phi^\ell} (1-\beta_{\phi^{\ell}}^{\ell}(s',o^{1:\ell})) \prod_{z=N-1}^{\ell+2} \beta_{\phi^z}^z(s',o^{1:z})}_{\text{ (7) $\ell+1$ terminates and $\ell$ does not}} + \\
\underbrace{ \sum_{i=1}^{\ell - 1}(1-\beta_{\phi^i}^i(s',o^{1:i})) \frac{ \partial Q_\Omega(s',o^{1:i})}{\partial \phi^\ell} \prod_{k=i+1}^{N-1} \beta_{\phi^k}^k(s',o^{1:k}) }_{\text{ (8) some relevant higher level options terminate}} \\
\end{split}
\end{equation}

Merging the first three terms as well as the 6th and 7th terms: 

\begin{equation} \label{workinggradient}
\begin{split}
\frac{\partial U(s',o^{1:\ell})}{\partial \phi^\ell} =  \prod_{j=N-1}^{\ell+1} \beta_{\phi^j}^j(s',o^{1:j}) \frac{\partial \beta_{\phi^\ell}^\ell(s',o^{1:\ell})}{\partial \phi^\ell}[ \underbrace{ -Q_\Omega(s',o^{1:\ell}) }_{\text{$\ell+1$ terminates and $\ell$ does not}} + \\
\underbrace{V_\Omega(s) [\prod_{j=\ell-1}^{1} \beta_{\phi^j}^j(s',o^{1:j})]}_{\text{all options terminate}}  +
\underbrace{ \sum_{i=1}^{\ell - 1}(1-\beta_{\phi^i}^i(s',o^{1:i})) Q_\Omega(s',o^{1:i}) [\prod_{k=i+1}^{\ell - 1} \beta_{\phi^k}^k(s',o^{1:k})]}_{\text{some relevant higher level options terminate}}  ] \\
+ \underbrace{(1-\beta_{\phi^{N-1}}^{N-1}(s',o^{1:N-1})) \frac{\partial Q_\Omega(s',o^{1:\ell})}{\partial \phi^\ell} }_{\text{none terminate}} 
+ \underbrace{ \frac{\partial V_\Omega(s)}{\partial \phi^\ell} \prod_{j=N-1}^{1} \beta_{\phi^j}^j(s',o^{1:j})}_{\text{all options terminate}}  + \\
\underbrace{ \frac{\partial Q_\Omega(s',o^{1:\ell})}{\partial \phi^\ell} \sum_{q=N-1}^{\ell+1} (1-\beta_{\phi^{q-1}}^{q-1}(s',o^{1:q-1})) \prod_{z=N-1}^q \beta_{\phi^z}^z(s',o^{1:z})}_{\text{only lower level options terminate}} + \\
\underbrace{ \sum_{i=1}^{\ell - 1}(1-\beta_{\phi^i}^i(s',o^{1:i})) \frac{ \partial Q_\Omega(s',o^{1:i})}{\partial \phi^\ell} \prod_{k=i+1}^{N-1} \beta_{\phi^k}^k(s',o^{1:k}) }_{\text{some relevant higher level options terminate}} \\
\end{split}
\end{equation}

We define the probability weighted advantage of not terminating $A_\Omega$ as:

\begin{equation} \label{Aw}
\begin{split}
A_\Omega(s',o^{1:\ell}) = Q_\Omega(s',o^{1:\ell}) - V_\Omega(s) [\prod_{j=\ell-1}^{1} \beta_{\phi^j}^j(s',o^{1:j})] - \sum_{i=1}^{\ell - 1}(1-\beta_{\phi^i}^i(s',o^{1:i})) Q_\Omega(s',o^{1:i}) [\prod_{k=i+1}^{\ell - 1} \beta_{\phi^k}^k(s',o^{1:k})]
\end{split}
\end{equation}

We proceed to substitute equation \eqref{Aw} into equation \eqref{workinggradient}: 

\begin{equation} 
\begin{split}
\frac{\partial U(s',o^{1:\ell})}{\partial \phi^\ell} =  - \prod_{j=N-1}^{\ell+1} \beta_{\phi^j}^j(s',o^{1:j}) \frac{\partial \beta_{\phi^\ell}^\ell(s',o^{1:\ell})}{\partial \phi^\ell} A_\Omega(s',o^{1:\ell}) \\
+ \underbrace{(1-\beta_{\phi^{N-1}}^{N-1}(s',o^{1:N-1})) \frac{\partial Q_\Omega(s',o^{1:\ell})}{\partial \phi^\ell} }_{\text{ (1) none terminate}} 
+ \underbrace{ \frac{\partial V_\Omega(s)}{\partial \phi^\ell} \prod_{j=N-1}^{1} \beta_{\phi^j}^j(s',o^{1:j})}_{\text{ (2) all options terminate}}  + \\
\underbrace{ \frac{\partial Q_\Omega(s',o^{1:\ell})}{\partial \phi^\ell} \sum_{q=N-1}^{\ell+1} (1-\beta_{\phi^{q-1}}^{q-1}(s',o^{1:q-1})) \prod_{z=N-1}^q \beta_{\phi^z}^z(s',o^{1:z})}_{\text{ (3) only lower level options terminate}} + \\
\underbrace{ \sum_{i=1}^{\ell - 1}(1-\beta_{\phi^i}^i(s',o^{1:i})) \frac{ \partial Q_\Omega(s',o^{1:i})}{\partial \phi^\ell} \prod_{k=i+1}^{N-1} \beta_{\phi^k}^k(s',o^{1:k}) }_{\text{ (4) some relevant higher level options terminate}}] \\
\end{split}
\end{equation}

Next we integrate out our last three terms so that they are in terms of a common derivative:

\begin{equation} 
\begin{split}
\frac{\partial U(s',o^{1:\ell})}{\partial \phi^\ell} =  - \prod_{j=N-1}^{\ell+1} \beta_{\phi^j}^j(s',o^{1:j}) \frac{\partial \beta_{\phi^\ell}^\ell(s',o^{1:\ell})}{\partial \phi^\ell} A_\Omega(s',o^{1:\ell}) \\
+ \underbrace{(1-\beta_{\phi^{N-1}}^{N-1}(s',o^{1:N-1})) \frac{\partial Q_\Omega(s',o^{1:\ell})}{\partial \phi^\ell} }_{\text{none terminate}} \\
+ \underbrace{ \prod_{j=N-1}^{1} \beta_{\phi^j}^j(s',o^{1:j}) \sum_{o'^{1}} ... \sum_{o'^{\ell}} \prod_{v=\ell}^{1} \pi_{\theta^v}^v(o'^v|s',o'^{1:v-1}) \frac{ \partial Q_\Omega(s',o'^{1:\ell})}{\partial \phi^\ell}}_{\text{all options terminate}}  + \\
\underbrace{ \frac{\partial Q_\Omega(s',o^{1:\ell})}{\partial \phi^\ell} \sum_{q=N-1}^{\ell+1} (1-\beta_{\phi^{q-1}}^{q-1}(s',o^{1:q-1})) \prod_{z=N-1}^q \beta_{\phi^z}^z(s',o^{1:z})}_{\text{only lower level options terminate}} + \\
\underbrace{ \sum_{i=1}^{\ell - 1}(1-\beta_{\phi^i}^i(s',o^{1:i})) \frac{ \partial Q_\Omega(s',o^{1:\ell})}{\partial \phi^\ell} \prod_{k=i+1}^{N-1} \beta_{\phi^k}^k(s',o^{1:k}) \prod_{p=i+1}^{\ell} \pi_{\theta^p}^p(o'^p|s',o'^{1:p-1})}_{\text{some relevant higher level options terminate}}] 
\end{split}
\end{equation}

We can then simplify the expression: 

\begin{equation} \label{initialpartial}
\begin{split}
\frac{\partial U(s',o^{1:\ell})}{\partial \phi^\ell} =  - \prod_{j=N-1}^{\ell+1} \beta_{\phi^j}^j(s',o^{1:j}) \frac{\partial \beta_{\phi^\ell}^\ell(s',o^{1:\ell})}{\partial \phi^\ell} A_\Omega(s',o^{1:\ell}) + \\
 [\underbrace{(1-\beta_{\phi^{N-1}}^{N-1}(s',o^{1:N-1})) \textbf{1}_{o'^{1:\ell}=o^{1:\ell}} }_{\text{none terminate}} 
+  \underbrace{ \prod_{j=N-1}^{1} \beta_{\phi^j}^j(s',o^{1:j}) \sum_{o'^{1}} ... \sum_{o'^{\ell}} \prod_{v=\ell}^{1} \pi_{\theta^v}^v(o'^v|s',o'^{1:v-1})}_{\text{all options terminate}}  + \\ 
\underbrace{ \sum_{q=N-1}^{\ell+1} (1-\beta_{\phi^{q-1}}^{q-1}(s',o^{1:q-1})) \prod_{z=N-1}^q \beta_{\phi^z}^z(s',o^{1:z}) \textbf{1}_{o'^{1:\ell}=o^{1:\ell}}}_{\text{only lower level options terminate}} + \\
\underbrace{ \sum_{i=1}^{\ell - 1}(1-\beta_{\phi^i}^i(s',o^{1:i})) \prod_{k=i+1}^{N-1} \beta_{\phi^k}^k(s',o^{1:k}) \prod_{p=i+1}^{\ell} \pi_{\theta^p}^p(o'^p|s',o'^{1:p-1})}_{\text{some relevant higher level options terminate}}] \frac{ \partial Q_\Omega(s',o'^{1:\ell})}{\partial \phi^\ell}\\
\end{split}
\end{equation}

We now note that substituting equation \eqref{simplify} into equation \eqref{simplederivative} yields:

\vspace{-5mm}
\begin{equation}
\frac{\partial Q_\Omega(s,o^{1:\ell})}{\partial \phi^\ell} = \gamma P(s'|s,o^{1:\ell}) \frac{\partial U(s',o^{1:\ell})}{\partial \phi^\ell}
\end{equation}
\vspace{-5mm} 

Substituting this expression into equation \eqref{initialpartial} we find that:

\begin{equation} 
\begin{split}
\frac{\partial U(s',o^{1:\ell})}{\partial \phi^\ell} = - \prod_{j=N-1}^{\ell+1} \beta_{\phi^j}^j(s',o^{1:j}) \frac{\partial \beta_{\phi^\ell}^\ell(s',o^{1:\ell})}{\partial \phi^\ell} A_\Omega(s',o^{1:\ell}) + \\
[ \underbrace{(1-\beta_{\phi^{N-1}}^{N-1}(s',o^{1:N-1})) \textbf{1}_{o'^{1:\ell}=o^{1:\ell}} }_{\text{none terminate}} + \underbrace{ \prod_{j=N-1}^{1} \beta_{\phi^j}^j(s',o^{1:j}) \sum_{o'^{1}} ... \sum_{o'^{\ell}} \prod_{v=\ell}^{1} \pi_{\theta^v}^v(o'^v|s',o'^{1:v-1})}_{\text{all options terminate}}  + \\ 
\underbrace{ \sum_{q=N-1}^{\ell+1} (1-\beta_{\phi^{q-1}}^{q-1}(s',o^{1:q-1})) \prod_{z=N-1}^q \beta_{\phi^z}^z(s',o^{1:z}) \textbf{1}_{o'^{1:\ell}=o^{1:\ell}}}_{\text{only lower level options terminate}} + \\
\underbrace{ \sum_{i=1}^{\ell - 1}(1-\beta_{\phi^i}^i(s',o^{1:i})) \prod_{k=i+1}^{N-1} \beta_{\phi^k}^k(s',o^{1:k}) \prod_{p=i+1}^{\ell} \pi_{\theta^p}^p(o'^p|s',o'^{1:p-1})}_{\text{some relevant higher level options terminate}}] \gamma P(s'|s,o^{1:\ell}) \frac{\partial U(s',o^{1:\ell})}{\partial \phi^\ell}\\
\end{split}
\end{equation}

Leveraging the augmented process structure and substituting in equation \eqref{T_PG}: 

\begin{equation}
\begin{split}
   \frac{\partial U(s',o^{1:\ell})}{\partial \phi^\ell} = - \prod_{i=\ell+1}^{N-1} \beta_{\phi^i}^i(s',o^{1:i}) \frac{\partial \beta_{\phi^\ell}^\ell(s',o^{1:\ell})}{\partial \phi^\ell} A_\Omega(s',o^{1:\ell}) + \sum_{s''} \sum_{o'^1} ... \sum_{o'^{\ell}} P_\gamma^{(1)}(s'',o'^{1:\ell}|s,o^{1:N-1}) \frac{ \partial U_\Omega(s'',o'^{1:\ell})}{\partial \phi^\ell} \\ 
   = - \sum_{s'',o'^{1:\ell}} \sum_{k=0}^\infty P_\gamma^{(k)}(s'',o'^{1:\ell}|s,o^{1:N-1}) \prod_{i=\ell+1}^{N-1} \beta_{\phi^i}^i(s',o^{1:i}) \frac{\partial \beta_{\phi^\ell}^\ell(s',o^{1:\ell})}{\partial \phi^\ell} A_\Omega(s',o^{1:\ell}),
  \end{split}
\end{equation}

We can then finally obtain that:

\begin{equation}
\begin{split}
\frac{\partial U(s_1,o_0^{1:\ell})}{\partial \phi^\ell} = -\sum_{s,o^{1:\ell}} \sum_{k=0}^\infty  P_\gamma^{(k)}(s,o^{1:\ell}|s_1,o_0^{1:N-1}) \prod_{i=\ell+1}^{N-1} \beta_{\phi^i}^i(s,o^{1:i}) \frac{\partial \beta_{\phi^\ell}^\ell(s,o^{1:\ell})}{\partial \phi^\ell} A_\Omega(s,o^{1:\ell}) \\
= -\sum_{s,o^{1:\ell}}  \mu_\Omega(s,o^{1:\ell}|s_1,o_0^{1:N-1}) \prod_{i=\ell+1}^{N-1} \beta_{\phi^i}^i(s,o^{1:i}) \frac{\partial \beta_{\phi^\ell}^\ell(s,o^{1:\ell})}{\partial \phi^\ell} A_\Omega(s,o^{1:\ell}).
\end{split}
\end{equation}

\section{Additional Details for Experiments} \label{ExpApp}

In Algorithm \ref{TabularAlg} we provide a detailed algorithm for our learning policy in the tabular setting. This algorithm generalizes the one presented in \citep{OC} for option-critic learning to hierarchical option-critic learning with $N$ levels of abstraction. In Algorithm \ref{AtariAlg} we provide the same generalization but from the Asynchronous Advantage Option-Critic model presented in \citep{Deliberation}. As in \citep{Deliberation} we use an $\epsilon$-soft policy leveraging the respective critic instead of learning a separate top level actor. As in \citep{OC} we potentially add in a regularization term $\eta$ for the termination policy update rule to decrease the likelihood that options terminate. In all of our experiments we used a discount factor of 0.99. 

\begin{algorithm}
  \caption{Hierarchical Option-Critic with Tabular Intra-option Q-Learning}\label{TabularAlg}
  \begin{algorithmic}
  \Procedure{LearnEpisode}{$env,N,\alpha,\gamma,\pi,\beta,\eta$}
  \State // get initial state
  \State $s \gets s_0$ 
  \State // select options for initial state
  \For{\texttt{$j = 1,...,N-1$}}
  \State $o^{j} \gets \pi^{j}(s,o^{1:j-1})$
  \EndFor
  \Repeat
  \State // take an action and step through the environment
  \State $a \gets \pi^N(a|s,o^{1:N-1})$
  \State $s^{'},r \gets env.step(a)$
  \State // calculate the expected discounted return 
  \State $r^{'} \gets r $
  \State \textbf{if} $s^{'}$ is non-terminal \textbf{then}
  \State\hspace{\algorithmicindent} $r^{'} \gets r^{'} + \gamma U(s',o^{1:N-1})$ (see equation \eqref{Ueq})
  \State // update the critic networks
  \For{\texttt{$j = 1,...,N-1$}}
  \State $\delta_j \gets r' - Q_U(s,o^{1:j})$
  \State $Q_U(s,o^{1:j}) \gets Q_U(s,o^{1:j}) + \alpha \delta_j$
  \EndFor
  \State $\delta_N \gets r' - Q_U(s,o^{1:N-1},a)$
  \State $Q_U(s,o^{1:N-1},a) \gets Q_U(s,o^{1:N-1},a) + \alpha \delta_N$
  \State // update the intra-option policies
  \For{\texttt{$j = 1,...,N-1$}}
  \State $\theta^j \gets \theta^j + \alpha_\theta \frac{\partial log \pi^j(o^j|s,o^{1:j-1})}{\partial \theta^j} Q_U(s,o^{1:j})$
  \EndFor
  \State $\theta^N \gets \theta^N + \alpha_\theta \frac{\partial log \pi^N(a|s,o^{1:N-1})}{\partial \theta^N} Q_U(s,o^{1:N-1},a)$
  \State // update the termination policies 
  \For{\texttt{$j = 1,...,N-1$}}
  \State $\phi^j \gets \phi^j - \alpha_\phi \prod_{i=j+1}^{N-1} \beta^i(s,o^{1:i}) \frac{\partial \beta^j(s,o^{1:j})}{\partial \phi^j}(A(s,o^{1:j})+\eta)$
  \EndFor
  \State // check which options have terminated and select new ones
  \State $o^{1:N-1} \gets chooseTerminatedOptions(s',o^{1:N-1},\pi,\beta, N)$
  \State // update the next state to now be the current state
  \State $s \gets s'$

  \Until{\text{$s^{'}$ is terminal}}
  \EndProcedure
  \Procedure{chooseTerminatedOptions}{$s,o^{1:k},\pi,\beta, k$} 
    
    \State \textbf{if} $\beta^k(s,o^{1:k}) = 1$
    \State\hspace{\algorithmicindent} \textbf{if} $k-1 = 1$
   \State\hspace{\algorithmicindent}\hspace{\algorithmicindent} $o^1 \gets \pi^{1}(s)$
    \State\hspace{\algorithmicindent} \textbf{else}
    \State\hspace{\algorithmicindent}\hspace{\algorithmicindent} $o^{1:k-1} \gets chooseTerminatedOptions(s,o^{1:k-1},\pi,\beta,k-1)$
    \State\hspace{\algorithmicindent} $o^k \gets \pi^{k-1}(s,o^{1:k-1})$

    \State \textbf{return} $o^{1:k}$
    \EndProcedure
  \end{algorithmic}
\end{algorithm}

\begin{algorithm}
  \caption{Asynchronous Advantage Hierarchical Option-Critic}\label{AtariAlg}
  \begin{algorithmic}
  \Procedure{LearnEpisode}{$env,N,\alpha,\gamma,\pi,\beta,\eta,T_{max},t_{min},t_{max}$}
  \State initialize global counter $T \gets 1$
  \State initialize thread counter $t \gets 1$
  \Repeat
  \State $t_{start} = t$
  \State $s_t \gets s_0$
  \State // reset gradients
  \State $dw \gets 0$ 
  \State $d\theta \gets 0$
  \State $d\phi \gets 0$
  \State // select options for initial state
  \For{\texttt{$j = 1,...,N-1$}}
  \State $o_t^{j} \gets \pi^{j}(s_t,o_t^{1:j-1})$
  \EndFor
  \Repeat
  \State // take an action and step through the environment
  \State $a_t \gets \pi^N(s_t,o_t^{1:N-1})$
  \State $s_{t+1},r_t \gets env.step(a_t)$
  \State // check which options have terminated and select new ones
  \State $o_t^{1:N-1} \gets chooseTerminatedOptions(s_{t+1},o_{t-1}^{1:N-1},\pi,\beta, N)$
  \State $t \gets t + 1$
  \State $T \gets T + 1$
  \Until{episode ends or $t - t_{start} == t_{max}$ or $(t - t_{start} > t_{min}$)          }
  \State $G = V(s_t)$ 
  \For{\texttt{$k = t-1,...,t_{start}$}}
  \State // accumulate thread specific gradients
  \State $G \gets r_k + \gamma G$ 
  \State // update the critic policies 
  \For{\texttt{$j = 1,...,N-1$}}
  \State $dw^j \gets dw^j + \alpha_w \frac{\partial (G - Q(s, o^{1:j}))^2}{\partial w^j}$
  \EndFor
  \State // update the intra-option policies
  \For{\texttt{$j = 1,...,N-1$}}
  \State $d\theta^j \gets d\theta^j + \alpha_\theta \frac{\partial log \pi^j(o^j|s,o^{1:j-1})}{\partial \theta^j} (G - Q(s,o^{1:j-1}))$
  \EndFor
  \State $d\theta^N \gets d\theta^N + \alpha_\theta \frac{\partial log \pi^N(a|s,o^{1:N-1})}{\partial \theta^N} (G - Q(s,o^{1:N-1}))$
  \State // update the termination policies 
  \For{\texttt{$j = 1,...,N-1$}}
  \State $d\phi^j \gets d\phi^j - \alpha_\phi \prod_{i=j+1}^{N-1} \beta^i(s,o^{1:i}) \frac{\partial \beta^j(s,o^{1:j})}{\partial \phi^j} (A(s,o^{1:j})+\eta)$
  \EndFor
  \EndFor
  \State update global parameters with thread gradients
  \Until{$T > T_{max}$}
  \EndProcedure
  \Procedure{chooseTerminatedOptions}{$s,o^{1:k},\pi,\beta, k$} 
    
    \State \textbf{if} $\beta^k(s,o^{1:k}) = 1$
    \State\hspace{\algorithmicindent} \textbf{if} $k-1 = 1$
   \State\hspace{\algorithmicindent}\hspace{\algorithmicindent} $o^1 \gets \pi^{1}(s)$
    \State\hspace{\algorithmicindent} \textbf{else}
    \State\hspace{\algorithmicindent}\hspace{\algorithmicindent} $o^{1:k-1} \gets chooseTerminatedOptions(s,o^{1:k-1},\pi,\beta,k-1)$
    \State\hspace{\algorithmicindent} $o^k \gets \pi^{k-1}(s,o^{1:k-1})$

    \State \textbf{return} $o^{1:k}$
    \EndProcedure
  \end{algorithmic}
\end{algorithm}

\subsection{Exploring four rooms}

\textbf{Hyperparameter search:} For the primitive actor-critic model our only tuned parameter is the learning rate over the range \{0.001,0.01,0.1,0.25,1.0,10.0\}. For the option-critic model we search over the number of options \{4,8,16\} and for the hierarchical option-critic model we use two options per layer of abstraction. All of our option models search over a intra-option learning rate shared among policies in the range \{0.01,0.1,0.5\}, a termination policy learning rate in the range \{0.01,0.1,0.25,1.0\} and a learning rate for critic models in the range \{0.1,0.5\}. 

\textbf{Selected hyperparameters:} For actor-critic learning we found it best to use a learning rate of 0.01, and a temperature of 0.1. For option-critic and hierarchical option critic learning we found it optimal to use a temperature of 1.0, a learning rate of 0.5 for the critics and intra-options policies, and a learning rate of 0.25 for the termination policies. It was best to use 4 options for option-critic learning. 

\textbf{Learning curve details:} We report the average number of steps taken in the last 100 episodes every 100 episodes, reporting the average of 50 runs with different random seeds for each algorithm.

\subsection{Discrete stochastic decision process}

\textbf{Hyperparameter search:} For the primitive actor-critic model our only tuned parameter is the learning rate over the range \{0.001,0.01,0.1,0.25,1.0,10.0\}. For the option-critic model we search over the number of options \{4,8,16\} and for the hierarchical option-critic model we use two options per layer of abstraction. All of our option models search over an intra-option learning rate shared among policies in the range \{0.01,0.1,0.5\}, a termination policy learning rate in the range \{0.01,0.1,0.25,1.0\} and a learning rate for critic models in the range \{0.1,0.5\}. 

\textbf{Selected hyperparameters:} A learning rate of 0.25 is used for actor-critic learning and the critics of the option architectures have a learning rate of 0.5. We found it beneficial to use higher temperatures with higher levels of abstraction using 0.01 for one level, 0.1 for two levels and 1.0 for three levels. For the option-critic architecture we found it optimal to use an intra-option learning rate of 0.1, and a termination learning rate of 0.01. For the hierarchical option-critic architecture we found it optimal to use an intra-option learning rate of 1.0, and a termination learning rate of 10.0. 4 options was best for the option-critic model. 

\textbf{Learning curve details:} We report the average reward over the last 100 episodes every 100 episodes, reporting the average of 10 runs with different random seeds for each algorithm.

\subsection{Multistory building navigation}

\textbf{Architecture details:} A core perceptual and contextualization model is shared across all policies and critics for each model to transform observations into conceptual states that can be processed to produce an option policy. The perceptual module was a 100 unit fully connected layer with ReLU activations. This perceptual module is processed by a 256 unit LSTM network with gradients truncated at 20 steps. Every intra-option policy, termination policy, and critic simply consists of one linear layer on top of this core module followed by a softmax in the case of intra-option policies and a sigmoid in the case of termination policies. 

\textbf{Hyperparameters:} We found optimal to use a learning rate of 1e-4 for all models a well as 16 parallel asynchronous threads and entropy regularization of 0.01 on the intra-option policies \citep{OC,Deliberation}

\textbf{Learning curve details:} We set our implementation of A3C to report recent learning performance after approximately 1 minute of training. Each minute we report the rolling mean reward calculated using a horizon of 0.99. To plot learning performance we take the average and standard deviation of the reported rewards over the past 1 million frames.  

\subsection{Atari multi-task learning}  \label{MTL}

\textbf{Experiment details:} In our Atari experiments we leverage the standard Open AI Gym v0 environments. A core perceptual and contextualization model is shared across all policies and critics for each model to transform observations into conceptual states that can be processed to produce primitive action and option policies. We follow architecture conventions for Atari games from \citep{A3C} to implement this module consisting of a convolutional layer with 16 filters of size 8x8 with stride 4,  followed by a convolutional layer with with 32 filters of size 4x4 with stride 2, followed by a fully connected layer with 256 hidden units. All three hidden layers were followed by a ReLU nonlinearity. This hidden representation is fed to a 256 unit LSTM network with gradients truncated at 20 steps.  Every intra-option policy, termination policy, and critic simply consists of one linear layer on top of this core module followed by a softmax in the case of intra-option policies and a sigmoid in the case of termination policies. The primitive action policy for each game is implemented with its own linear layer followed by a softmax as the games have different action spaces. In our experiments on Atari we followed conventions from past work using 16 parallel asynchronous threads and entropy regularization of 0.01 on the intra-option policies \citep{OC,Deliberation}. We use a learning rate of 1e-4 for each model.

\textbf{Analysis of learned options for multi-task learning:} In Table \ref{switching} we detail the average option switching frequencies for each of the 21 Atari games when we train in a many task learning setting. For the option-critic architecture and three-level hierarchical option-critic architecture we define a switch as terminating an option at a particular level and choosing a new different option at that level. We can see that the hierarchical option-critic architecture displays much greater variation in its option switching frequencies across games.  

\begin{table}
\centering
\begin{tabular}{c|c|cc}
\toprule
Environment  & OC & HOC ($o^1$) & HOC ($o^2$) \\ \hline
Alien & 5.4 & 7.7 & 1.5 \\ 
Amidar & 5.5 & 6.5 & 1.7 \\ 
Assault & 4.0 & 3.3 & 1.9 \\ 
Atlantis & 5.3 & 6.9 & 1.7 \\ 
BankHeist & 5.5 & 7.8 & 2.6 \\ 
BattleZone & 5.0 & 6.6 & 1.8 \\ 
BeamRider & 5.4 & 3.2 & 1.8 \\ 
Berzerk & 5.4 & 6.6 & 1.9 \\ 
Carnival & 5.5 & 4.4 & 2.0 \\ 
Centipede & 4.3 & 6.7 & 3.1 \\ 
ChopperCommand & 5.5 & 6.3 & 1.6 \\ 
DemonAttack & 5.4 & 3.4 & 1.7 \\ 
Jamesbond & 4.8 & 6.5 & 1.7 \\ 
MsPacman & 5.5 & 7.6 & 7.5 \\ 
Phoenix & 4.5 & 3.2 & 1.9 \\ 
Riverraid & 5.0 & 7.7 & 1.5 \\ 
Solaris & 3.4 & 5.6 & 2.7 \\ 
SpaceInvaders & 4.1 & 6.0 & 2.6 \\ 
Tutankham & 5.2 & 9.7 & 7.8 \\ 
WizardOfWor & 3.9 & 9.1 & 2.2 \\ 
Zaxxon & 5.5 & 4.2 & 1.7 \\ 
\bottomrule
\end{tabular}
\caption{The average number of steps before switching options by game for the median performance option-critic (OC) and hierarchical option-critic (HOC) architectures during the evaluation period. For our three level model, we detail statistics for high level option $o^1$ as well as low level option $o^2$.} \label{switching}
\end{table} 

\begin{figure}
	\includegraphics[scale=0.145]{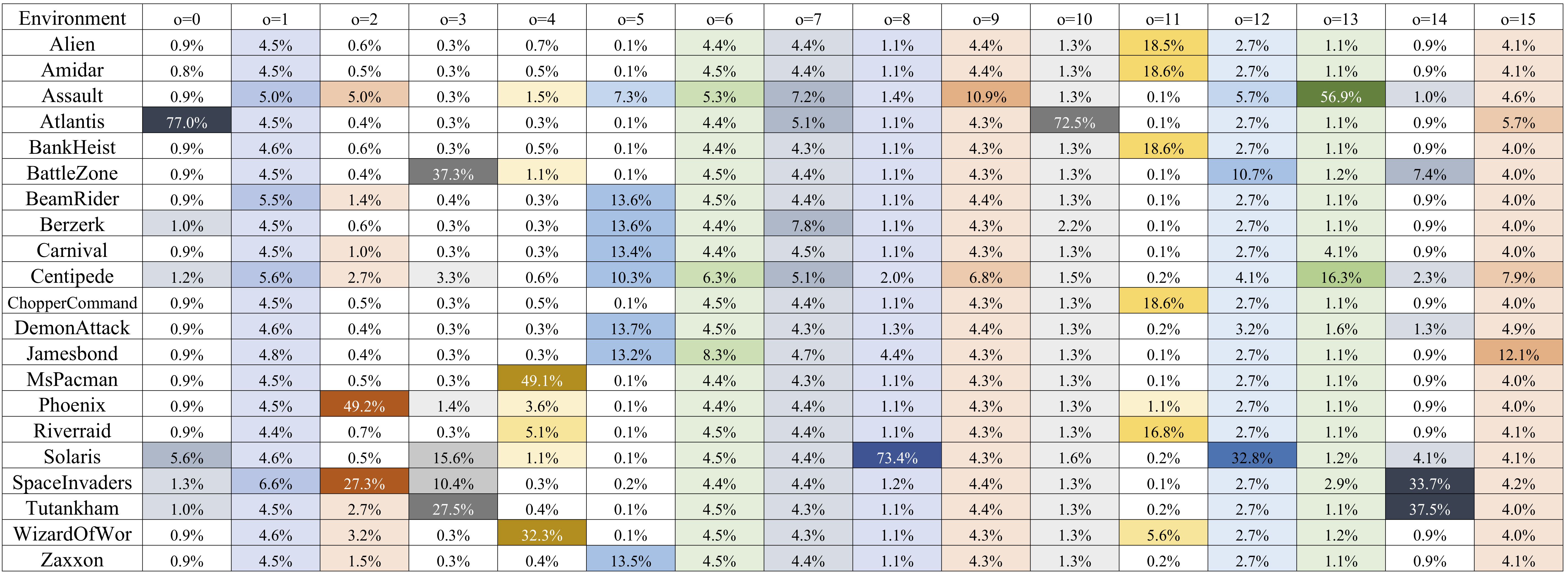}
	\vspace{-4mm}
\caption{Option specialization across Atari games for a 16 option Option-Critic architecture trained in the many task learning setting.}
	\label{OCFull}
	\vspace{-3mm}
\end{figure}

\textbf{Details on figures analyzing options:} In the main text we provide option specialization across Atari games for all 9 possible option combinations for the hierarchical option-critic architecture and the top 9 most used options for the option-critic architecture to save space. In Figure \ref{OCFull} we provide detailed information including the specialization of all learned options for the option-critic architecture. In all of our option analysis figures we use a heat-map where each option is assigned a color. This way options can be clearly separated from the surrounding options on the grid. We keep cells for options that are used on a game less than 1\% of the time white. We then add a light color that gets progressively darker at 5\% specialization, 10\% specialization, and 25\% specialization. 

\subsection{Comparison with methods for multi-task and lifelong learning}

In this work we explore a relatively straightforward application of multi-task learning on the Atari games. Following conventions in multi-task learning \citep{Caruana97}, as the action space is different with varying sizes across games, all parts of the network are shared with the exception of a task specific layer in the last layer of the policy over primitive actions. This a somewhat arbitrary choice of the extent of weight sharing in light of recent work that focuses on more dynamic sharing patterns in multi-task learning, lifelong learning, and continual learning settings \citep{HLS,Cross,RoutingNets,PNN,PathNets,MTD,Kirkpatrick17,LwF,FC,GenerativeDist}. A more dynamic weight sharing pattern should allow the hierarchical option-critic architecture to potentially achieve better sample efficiency in a multi-task learning setting. However, we leave analysis of the proper way to achieve this in a general sense to future work as it is largely orthogonal to our main contribution of presenting policy gradient theorems to optimize a deep hierarchy of options.

Our approach is also orthogonal to recent approaches improving the efficiency of multi-task learning through a learned curriculum learning process \citep{MTL,Graves18}. In the setting we explore, all models train on the games in a balanced fashion throughout time and the agent is not assumed to have any control over which environment it trains on. Controlling the curriculum of games to train on could also potentially improve the efficacy of our approach. 

\end{document}